\documentclass[sigconf,nonacm]{acmart}  

\renewcommand\footnotetextcopyrightpermission[1]{}
\copyrightyear{2025}      

\acmYear{}                
\acmVolume{}              
\acmNumber{}              
\acmDOI{}                 
\acmISBN{}                
\acmConference{}          

\usepackage{blindtext} 
\usepackage{float}
\usepackage{xcolor}
\usepackage{amsmath}
\usepackage{booktabs}
\usepackage{multirow} 
\usepackage{siunitx} 
\usepackage{adjustbox}
\usepackage{array}
\usepackage[ruled,vlined]{algorithm2e}
\usepackage{enumitem}
\usepackage{bm}

\newcommand{\point}[1]{\par\smallskip{\noindent\textbf{#1.}~}}
\newcommand{\colonpoint}[1]{\noindent\textbf{#1:}~}
\newcommand{\Name}{\textsf{TeleSparse}}

\newtheorem{theorem}{Theorem}[section]

\newcommand{\poly}{\ensuremath{\mathsf{poly}}\xspace}
\newcommand{\negl}{\ensuremath{\nu}\xspace}

\begin{document}

\title[TeleSparse: Practical Privacy-Preserving Verification of Deep Neural Networks]{TeleSparse: Practical Privacy-Preserving Verification of Deep Neural Networks}

\author{Mohammad M Maheri}
\affiliation{%
  \institution{Imperial College London}
  \country{}
}

\author{Hamed Haddadi}
\affiliation{%
  \institution{Imperial College London}
  \country{}
}

\author{Alex Davidson}
\affiliation{%
  \institution{LASIGE, Universidade de Lisboa}
  \country{}
}

\begin{abstract}
Verification of the integrity of deep learning inference is crucial for understanding whether a model is being applied correctly. However, such verification typically requires access to model weights and (potentially sensitive or private) training data. So-called Zero-knowledge Succinct Non-Interactive Arguments of Knowledge (ZK-SNARKs) would appear to provide the capability to verify model inference without access to such sensitive data. However, applying ZK-SNARKs to modern neural networks, such as transformers and large vision models, introduces significant computational overhead.

We present \Name, a ZK-friendly post-processing mechanisms to produce practical solutions to this problem. \Name\ tackles two fundamental challenges inherent in applying ZK-SNARKs to modern neural networks:
(1) Reducing circuit constraints: Over-parameterized models result in numerous constraints for ZK-SNARK verification, driving up memory and proof generation costs. We address this by applying \emph{sparsification} to neural network models, enhancing proof efficiency without compromising accuracy or security.
(2) Minimizing the size of lookup tables required for non-linear functions, by optimizing activation ranges through neural \emph{teleportation}, a novel adaptation for narrowing activation functions' range.

\Name\ reduces prover memory usage by 67\% and proof generation time by 46\% on the same model, with an accuracy trade-off of approximately 1\%.
We implement our framework using the Halo2 proving system and demonstrate its effectiveness across multiple architectures (Vision-transformer, ResNet, MobileNet) and datasets (ImageNet,CIFAR-10,CIFAR-100). This work opens new directions for ZK-friendly model design, moving toward scalable, resource-efficient verifiable deep learning.
\end{abstract}



\maketitle

\section{Introduction}
\label{sec:introduction}
Deep learning models have had considerable success across various machine learning (ML) tasks in recent years. The proliferation of large-scale deep learning models, characterized by their parameter-intensiveness (i.e. in the millions), has become commonplace. However, as the number of parameters in these models increases, several challenges pertaining to trustworthiness in ML arise.

\begin{enumerate}
    \item Due to the escalating number of parameters and computational demands for training or inferring the model, consumers often opt to outsource these computations to service providers, a practice known as ML-as-a-service (MLaaS). Consequently, ensuring the integrity of predictions becomes crucial, especially in cases where consumers lack trust in the reliability of the model provider.
    \item Training large deep neural networks (DNN) often involves vast amounts of privacy-sensitive or proprietary data, and demands substantial computational resources. As a result, the trained model becomes valuable intellectual property for the trainer. Publishing model weights for auditing purposes~\cite{xue2020auditing} is typically impractical. This creates a need for methods to verify specific model properties, while keeping the weights private from the verifier.

    \item Where client data is highly sensitive, model providers may opt to send the model to client devices for local inference (e.g., large language models). However, this introduces the risk of intentional or unintentional deviations from correct computations. Such deviations may result from attacks on vulnerable client devices or malicious clients attempting to benefit from producing incorrect model predictions.
\end{enumerate}

Consequently, ensuring the correctness of model inference becomes paramount for model providers or any other third-party seeking to trust the model's output predictions. In all three above scenarios, the integrity of model inference is critical, necessitating the implementation of verifiable model inference mechanisms.

\point{ZK-SNARK}
Verifying the integrity of inference while preserving privacy of model weight is a critical challenge, as previous research has shown that revealing model weights can expose privacy of the training data through attacks like membership inference~\cite{leino2020stolen,Cretu2024,wang2024gcl}. Cryptographic primitives, particularly zero-knowledge succinct non-interactive arguments of knowledge (ZK-SNARKs)~\cite{ITCS:BCCT12,kilian1992note,micali2000computationally}, have emerged as a solution to verifying inference outputs in a privacy-preserving manner~\cite{ben2014succinct,liu2021zkcnn,lee2024vcnn,weng2023pvcnn,kang2022scaling}.
Most machine learning applications require succinctness in zero-knowledge proofs (ZKP), especially when verifiers operate on resource-constrained devices like edge or mobile platforms. While recent ZK constructions, such as those proposed in~\cite{li2023zksql}, aim to reduce the high prover costs associated with ZK-SNARKs, they often lack the compact proof size essential for ML applications. 
The SNARK-based approach holds promise due to: i) its ability to generate succinct proofs, typically verifiable in under 1 second; and ii) the verification process can be conducted by any third party solely by accessing the generated proof, demonstrating public verifiability and eliminating the need for interaction despite methods based on interactive ZK protocol~\cite{weng2021mystique}. While ZK-SNARK presents a promising avenue for verifiable deep models, it necessitates significant computational resources on the prover's side, limiting its application primarily to toy example deep learning models. To mitigate this challenge, various cryptographic techniques have been explored, including: i) quantization NN to low precision integers (8bit or 16bit)~\cite{feng2021zen}; ii) utilization of lookup tables for non-linear functions~\cite{kang2022scaling}; and iii) efficient representation of relations for convolution equations~\cite{weng2023pvcnn,lee2024vcnn,liu2021zkcnn} and attention mechanism in transformer architecture~\cite{sun2024zkllm}. However, there has been limited exploration into post-processing deep machine learning models to make them ZK-friendly, in order to reduce the required resources for proof verification. 

\point{Computational overheads} 
With the rise of parameter-intensive neural networks, large-scale models are increasingly employed in various ML applications, such as vision and language models. The cryptographic optimizations mentioned can decrease memory usage and proof generation time for ensuring the integrity of model inferences. However, with significant overheads to naively composing such proofs with such models still present, it is necessary to develop innovative approaches beyond simply using ZK-SNARKs. 

Inspired by the emerging trend of tiny ML~\cite{lin2023tiny,zaidi2022unlocking}, compression of models appears promising in this regard. 
However, such techniques should possess specific properties. 
Firstly, they should be lightweight, ensuring that model compression requires less computation than the reduction of resources needed for proof generation, resulting from said compression.
Secondly, compressing the model weights and/or activation values should preserve its performance as much as possible, in terms of maintaining its overall accuracy. If compression compromises model performance, the proof generation step becomes futile.
Finally, compression techniques should be adaptable to zero-knowledge proving systems, reducing the required resources without compromising the security guarantees of the proving system.

\point{Our work}
In this study, we investigate the root causes of proof generation overhead in ZK-SNARK systems (both in terms of memory usage and computational time) when applied to modern deep learning architectures, such as transformers~\cite{vaswani2017attention}. We identify two primary challenges. First, the growing complexity of modern architectures like transformers, which contain significantly more parameters than earlier models, results in a corresponding increase in the number of constraints required by ZK-SNARK backends for verification. Second, handling non-linear functions in DNNs poses a challenge for ZK systems, as these functions cannot be directly encoded in arithmetic circuits. In sum-check based proving systems, this limitation is addressed by using bit decomposition and polynomial approximations of non-linear functions~\cite{liu2021zkcnn, ghodsi2017safetynets, weng2023pvcnn}. However, these methods introduce computational overhead and can reduce model accuracy due to the inherent approximation of complex functions~\cite{feng2021zen, sun2024zkllm}. In particular, polynomial approximations are not fully precise representations of the original functions.
Recent advances in verifiable machine learning models~\cite{sun2024zkllm,kang2022scaling}~---~utilizing lookup tables to handle the non-linear activation functions used in DNNs~---~still face significant challenges, due to the need for large lookup tables to cover the broad input ranges of activations. These lookup tables introduce considerable overhead, leading to slower proof generation, increased memory demands for the prover, and larger proof sizes. This challenge is particularly acute in transformer architectures, which are extensively employed in large language models (LLMs)~\cite{xu2022systematic,radford2018improving} and modern vision models such as Vision Transformers (ViTs)~\cite{dosovitskiy2020image}. These models are known to produce outlier activation values~\cite{ashkboos2024quarot,bondarenko2023quantizable,sun2024massive}, exacerbating the issues associated with lookup table size and efficiency.
 
To address the first problem, we carefully adopt a lightweight pruning technique that suits the constraints and requirements of ZK-SNARK verification, in a resource-efficient manner. This approach reduces the number of constraints necessary for circuit verification by \emph{sparsifying} the model. In other words, allowing inference verification on a pruned model that meets both computational and memory efficiency demands essential for zero-knowledge proofs. 

To address the second, we identify symmetric neural network configurations that minimize the range of activation values, thereby reducing the size of the necessary lookup tables. This approach significantly decreases the computational and memory overhead associated with proof generation: a crucial step toward more efficient and scalable ZK-SNARK verification for modern neural network architectures. We achieve this symmetry discovery by formulating an optimization problem over neural teleportation~\cite{armenta2023neural}, a technique initially developed to accelerate and study neural network training, here adapted to optimize the activation input range. This novel application of neural teleportation allows us to efficiently manage activation ranges in models, maintaining both verification speed and resource efficiency in zero-knowledge contexts.

To implement ZK-SNARKs under the outlined constraints, we leverage Halo2~\cite{zcash-halo2}, a recent proving system that provides a highly efficient and flexible backend for ZK applications. Halo2 is a ZK-SNARK constructions that builds on the principle of recursive proof composition and operates without a trusted setup~\cite{bowe2019recursive}, making it particularly suitable for layer-wise DNNs. 
Moreover, its flexible arithmetization with lookup argument capabilities~\cite{gabizon2019plonk} handles non-linear DNN functions efficiently.
Building on this framework, \Name\ achieves about 67\% reduction in proof generation time and a 46\% reduction in prover memory usage, with a minimal ~1\% performance loss on the CIFAR-100~\cite{krizhevsky2009learning} dataset. This carefully chosen pruning strategy is post-processed, making it compatible with any pre-trained transformer model without requiring modifications to the training pipeline and adding less than 1\% memory overhead.

\point{Contributions}
We present the following key contributions:
\begin{itemize}

    \item We bridge the gap in adapting modern DNN architectures like transformers to ZK-SNARKs, by post-processing pre-trained models for scalability. This approach identifies two key factors to reduce overheads: minimizing circuit constraints in ML models, and optimizing lookup table ranges.
    
    \item We introduce model sparsification to efficiently generate ZK-SNARK proofs, specifically by devising a method that leverages sparsity to reduce the number of circuit constraints. This approach not only enhances proof efficiency but also maintains the security of the underlying proving system, which we formalize in our methodology.

    \item We present a novel application of neural teleportation to optimize activation ranges, reducing lookup table sizes. This approach, originally used for improving training convergence, proves effective across a wide range of models, including transformer-based architectures.

    \item We implement \Name\ on diverse architectures (Vision Transformers, ResNet, MobileNet) and datasets (CIFAR-10, CIFAR-100, ImageNet), demonstrating its efficiency,  scalability, and accuracy compared to state-of-the-art methods.
\end{itemize}

Overall, our contributions are orthogonal to existing ZK-SNARK research that focuses on optimizing the arithmetization of DNN layers (like CNNs and transformers)~\cite{liu2021zkcnn,weng2021mystique} and parallelizing proof generation using GPU resources~\cite{sun2024zkllm}. Instead, our work explores how to make DNN models themselves ZK-friendly—specifically adapting models to be computationally efficient and aware of ZK system requirements. This approach offering a synergistic pathway toward scalable and ZK-efficient ML.

On top of our stated results, we believe that our work can open a new path for designing DNN training or post-training, to make the output model more ZK-friendly. The code is available at \href{https://github.com/mammadmaheri7/TeleSparseRepo}{this link}. 

\section{Related Works}

\subsection{Verifiable deep model inference}
\label{subsection:verifiable_dnn_inference}


The field of secure inference in machine learning (ML) has seen rapid growth in protecting various aspects: i) privacy of inputs, ii) privacy of model parameters, and iii) integrity of model computation against potential malicious actors. Prior efforts to address these critical issues have employed techniques such as multi-party computation (MPC), homomorphic encryption (HE), or zero-knowledge (ZK) proofs. MPC methods distribute computation across multiple parties to prevent the exposure of inputs (e.g., model weights and ML model's input)~\cite{knott2021crypten,kumar2020cryptflow,srinivasan2019delphi,zeng2023mpcvit}. However, these approaches require synchronous interaction among parties, which is often impractical in many ML scenarios. Additionally, most MPC protocols only consider semi-honest adversaries~\cite{knott2021crypten,kumar2020cryptflow,srinivasan2019delphi,zeng2023mpcvit}, making them susceptible to parties deviating from the protocol.
Another approach involves leveraging HE, which allows computation on encrypted data. 
However, HE does not provide verification for the integrity of ML computations. Moreover, HE entails high computational costs, rendering it impractical for deep models, particularly parameter-intensive models~\cite{lou2021hemet,lee2022low}.
As the demand for publicly verifiable proof in ML computations rises in various scenarios, efforts have pivoted towards generating ZK proof of ML computations.

Recent research has explored using ZK proving systems to introduce verifiability in the inference of machine learning models. An initial study~\cite{feng2021zen} utilized Groth16~\cite{groth2016size} for verifying neural network inference.
To mitigate the significant overhead of proof generation, several studies have introduced novel computation representations, such as Quadratic polynomial programs \cite{lee2024vcnn,kosba2014trueset}. These representations are tailored to model-specific architectures, particularly Convolutional Neural Networks (CNNs)~\cite{liu2021zkcnn,lee2024vcnn,fan2024validcnn,balbas2023modular} and Transformers~\cite{sun2024zkllm}. 
These approaches are constrained by their focus on a specific model architecture, and most of them~\cite{liu2021zkcnn,lee2024vcnn,fan2024validcnn,feng2021zen,balbas2023modular} struggle with inefficiency in handling non-linear functions (e.g., common activation functions like ReLU), due to the limitations imposed by the underlying constraint systems of their proving systems. They introduce additional overhead because of the need for bit-decomposition or approximating activation functions with polynomial representations, both of which are ineffective for modern activation functions like Gaussian error linear unit (Gelu)~\cite{hendrycks2016gaussian}.
To address the non-linearly functions challenge,~\cite{kang2022scaling} employed the more recent proving system Halo2~\cite{zcash-halo2} and adopted Plonk arithmetization~\cite{gabizon2019plonk} to utilize lookup tables for non-linear functions.
Alternative lookup proofs~\cite{hao2024scalable,lu2024efficient} reduce computational costs but differ from PLONK arithmetization.
Specifically, they do not ensure constant proof size (succinctness) or maintain non-interactivity, which would allow for minimal communication between the prover and verifier.
As noted, previous research efforts attempted to represent neural network model computations efficiently based on the constraint systems provided by their proving mechanisms.
Although they have investigated efficient ZK proving by employing cryptographic techniques, they overlooked making DNNs ZK-friendly.
To address this gap, we propose post-training on the deep model to reduce both proving time and memory consumption significantly.

\subsection{Sparse training/inference}
Prior research~\cite{narang2017exploring, frankle2018lottery, jayakumar2020top} has shown that a small subset of a fully trained DNN is sufficient to represent the learned function. This property of DNNs allows for the development of efficient architectures, reducing computational and memory requirements during training (sparse training)~\cite{jayakumar2020top, sung2021training, liu2021we, raihan2020sparse, zhang2023bi, lasby2023dynamic} or through post-training techniques~\cite{kurtic2024ziplm, sun2023simple, kwon2022fast, ma2023llm}. Sparse training methods focus on decreasing computational load and model footprint during training, requiring optimized algorithms from the early stages of the training process. On the other hand, post-training pruning techniques prune unnecessary weights from a well-optimized model (dense model) using calibration data. These methods aim to reduce inference time and model size while preserving the model's performance.

Pruning techniques are typically classified into structured and unstructured methods. Structured pruning~\cite{ma2023llm, fang2023depgraph, yu2023x} generally removes entire neurons, channels, or filters, which can lead to a significant drop in accuracy and often requires extensive fine-tuning to recover performance. In contrast, unstructured pruning~\cite{kuznedelev2024cap, xu2023efficient, bai2024gradient} targets the removal of individual weights, better preserving the model's performance. However, unstructured pruning requires careful hardware implementation to reduce inference delay effectively.

Although the benefits of sparsification for reducing model size and improving hardware efficiency have been explored, sparse neural networks have not yet been utilized for efficient ZK-proof model verification. In our work, we propose a method to leverage sparsification to reduce the overhead of ZK-SNARK proof generation. Additionally, \Name\ supports unstructured sparsification, crucial for maintaining the model's performance. To the best of our knowledge, this is the first study to explore pruning methods that specifically reduce memory consumption and CPU computation while generation ZK-SNARK proofs for sparse DNN models.

\subsection{Neural Network Teleportation}
Continuous symmetries in neural networks refer to the invariance in the parameter space, where certain transformations of the network's weights do not affect its output. These symmetries emerge as a result of overparameterization, where multiple distinct weight sets can represent the same model function~\cite{gluch2021noether}. Such symmetries have been observed in networks with homogeneous activation functions~\cite{badrinarayanan2015symmetry, du2018algorithmic}, as well as in other architectural components, such as softmax and batch normalization~\cite{kunin2020neural}. Investigating these symmetries has contributed to improved training optimization and better generalization.

Neural teleportation, explored through quiver representation theory~\cite{armenta2021representation}, exploits symmetries in the loss landscape. It moves network parameters to a different point with the same objective value, enabling faster convergence in gradient-based optimization. This helps traverse the loss landscape efficiently by moving between different symmetric configurations of the model~\cite{armenta2023neural}.

Several works have expanded on neural teleportation and continuous symmetries.
For example,~\cite{zhao2022symmetry} proposed a symmetry teleportation algorithm that not only searches for optimal teleportation destinations but also leverages symmetries to maximize the gradient, thereby accelerating gradient descent. Additionally, they developed a general framework based on equivariance to analyze the loss landscape and the dimensions of minima induced by symmetries. 

In this work, we propose a novel application of neural teleportation to minimize the range of inputs to activation functions. This approach effectively mitigates outlier activation inputs~\cite{bondarenko2023quantizable}, which leads to large lookup tables in ZK proof generation. Furthermore, we extend the concept of neural teleportation to modern neural networks that use non-scale-invariant activation functions, such as GELU, which have been less explored in prior research.

\section{System and Threat Model}

\point{Notation}
The mathematical notations are listed in Appendix~\ref{sec:appendix_notation_table}.

The goal of our system, shown in Figure~\ref{fig:overview_zksnark}, is to ensure verifiable and privacy-preserving inference in DNN models. The system involves two primary parties:

\begin{figure}
    \centering
    \includegraphics[width=0.8\linewidth]{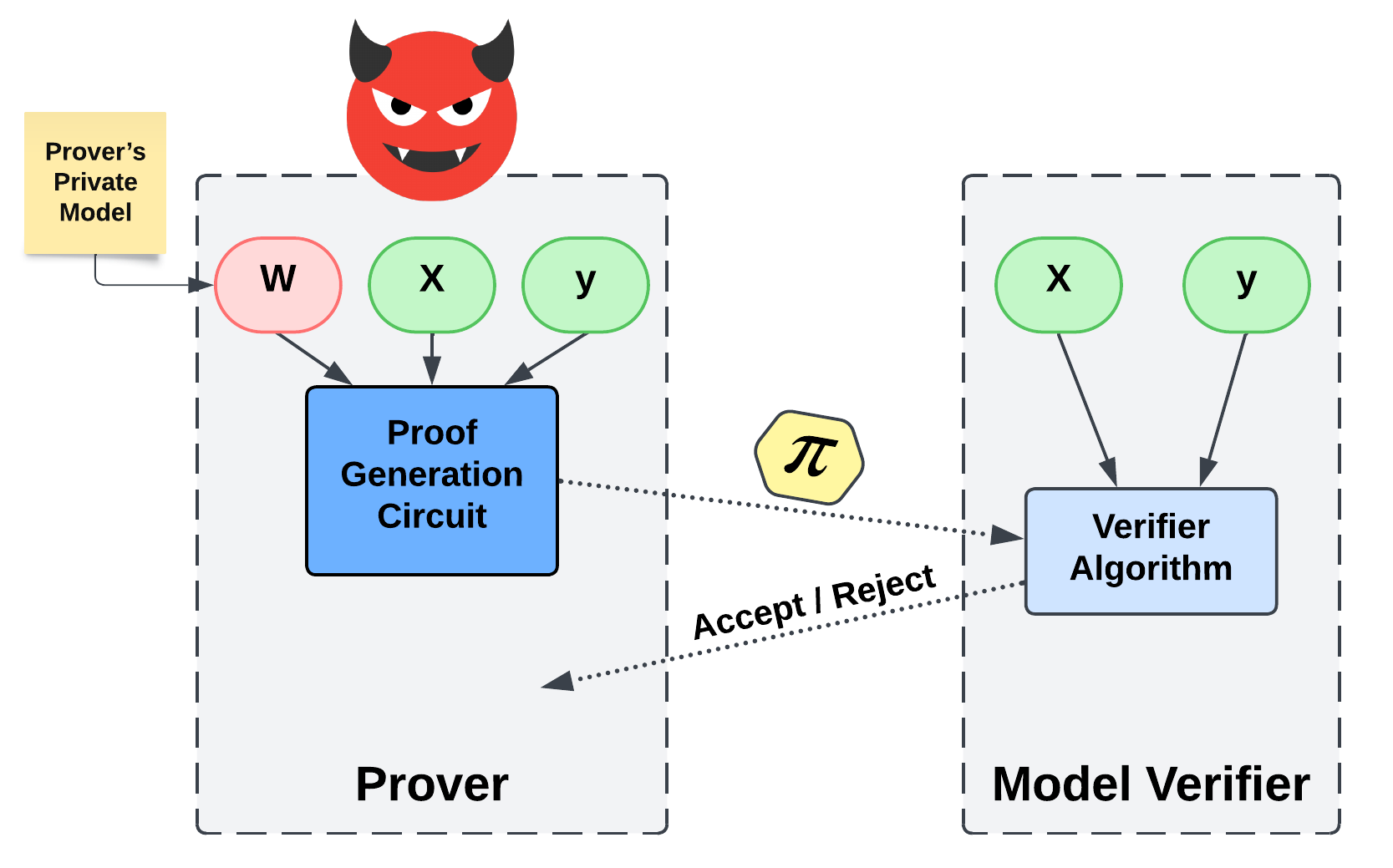}
    \caption{System diagram of ZK-SNARK DNN inference.}
    \label{fig:overview_zksnark}
\end{figure}

\begin{enumerate}
    \item Prover ($P$): Owns a private deep learning model with parameters ($W$). The prover computes the model's output ($y$) for a given input ($X$) and generates a proof ($\pi$) to demonstrate the correctness of the computation, i.e., $y = f(X; W)$, securely and privately.
    \item Verifier ($V$): Receives the proof ($\pi$) and verifies that $y = f(X; W)$ without learning anything about $W$, ensuring the integrity and confidentiality of the model.
\end{enumerate}
The private deep learning model inference can be represented as $y = f(X; W)$, where:

\begin{itemize}
\item $X$: Public input to the ZK circuit, representing the input data to the DNN.
\item $y$: Public input to the ZK circuit, corresponding to the output of the DNN computation.
\item $W$: Private input to the ZK circuit, representing the parameters (weights) of the DNN.
\end{itemize}

In pursuit of this objective, we utilize ZK-SNARK~\cite{ITCS:BCCT12}, a cryptographic protocol that enables Prover $P$ to generate a proof $\pi$. This proof enables a Verifier $V$ to ascertain, with only the knowledge of $\pi$, $y$, and $X$, that the Prover possesses certain parameters $W$ satisfying $y = f(X; W)$.
Following the Halo2 protocol, which is an instantiation of a ZK-SNARK proving system, our system inherits all security properties of Halo2~\cite{ITCS:BCCT12,kang2022scaling}, 
including Knowledge Soundness, Zero-knowledge detailed in Appendix~\ref{sec:appendix_halo2_properties}.

The proving system operates under the security assumption that adversaries are computationally limited~\cite{bunz2020transparent}. Even if adversaries deviate from the proving protocol, they cannot generate a valid proof for an incorrect computation. This ensures that malicious adversaries cannot compromise privacy (zero-knowledge property) or undermine the knowledge soundness of the ZK-proving system. We discuss the potential threats arising within our system and security model below in Section~\ref{sec:potential-threats}. Regardless, note that this represents a stronger security assumption compared to the honest-but-curious threat model commonly adopted in differential privacy (DP) and multi-party computation (MPC) methodologies.

As in prior works~\cite{kang2022scaling,weng2023pvcnn,groth2016size}, we assume that the model architecture is public (known to the verifier), while the model weights remain private. This assumption aligns with the trend toward open-source models~\cite{kang2022scaling}. Such an approach is particularly relevant for practical applications where model providers aim to protect the intellectual property of their weights  while enabling the verification of computations. Applications are detailed in Appendix~\ref{sec:appendix_applications}.

\subsection{Potential Threats}
\label{sec:potential-threats}
The system addresses the following threats:
\begin{itemize}
\item Privacy Threats: Ensures that $W$ (the model's parameters) remains private and that $V$ learns nothing beyond $y$. 
However, since the verifier observes the proof, an adversary could attempt to extract information about $W$ from it, posing a potential privacy risk. We analyze this leakage and its implications in Section \ref{sec:privacyAnalysis}.
\item Integrity Threats: Prevents adversaries from generating valid proofs (accepted by $V$) for incorrect computations of $f$ or tampering with $X$ or $y$.
\end{itemize}

\subsection{Design Goals}
\label{sec:design_goal}
The primary objective of our approach is to make ZK-SNARKs practical and scalable for verifiable inference on modern DNNs.
As noted in Section~\ref{subsection:verifiable_dnn_inference}, ZK-SNARKs enable efficient verifiable computation due to their ability to generate succinct proofs, which drastically reduces the verifier's computational load. This efficiency, however, comes at the cost of substantial prover overhead in terms of required memory and proof generation time. For instance, proving even a relatively small model, like the Tiny Vision Transformer (Tiny-ViT)~\cite{dosovitskiy2020image}, requires over 10TB of memory, which exceed the capacity of most practical systems.

To overcome the challenges associated with ZK-SNARKs for DNNs, we have outlined several key objectives that collectively enable practical and scalable verifiable inference on modern DNNs:
\begin{enumerate}

    \item Reducing the Number of Circuit Constraints ($G_1$): Modern DNNs are typically parameter-intensive, resulting in an enormous number of operations, often in the billions for models such as vision transformers. To construct a ZK-SNARK, each operation within the neural network must be “arithmetized,” or converted into circuit constraints, which the prover must satisfy to generate a valid proof. However, as the number of operations—and consequently, the number of constraints—increases, so do the memory usage and proof generation time required by the prover. Our approach addresses this by leveraging neural network sparsification to reduce the number of constraints associated with DNNs. Specifically, we devise a way to efficiently prove statements about sparse DNNs, preserving predictive performance while significantly reducing prover memory and computation costs. Detailed implementation of this approach is covered in Section~\ref{sec:sparsification}.

    \item Reducing Lookup Table Argument Overhead ($G_2$): Non-linear functions commonly used in DNNs, such as activation functions, present a significant challenge in ZK proof generation. These functions often need to be approximated through large lookup tables to fit within a ZK circuit framework. As prior work has shown, non-linear layers in models like ResNet-101 can consume up to 80\% of the prover's computational load due to extensive lookup table operations~\cite{weng2021mystique,hao2024scalable}. Although Halo2 provides efficient support for lookup arguments, our analysis reveals that non-linear functions still account for a substantial portion of the prover's computation and increase proof size. To address this, we propose an optimized neural network teleportation detailed in Section~\ref{sec:teleportation}.

\end{enumerate}

\subsection{Methodology Overview}

\begin{figure}
    \centering
    \includegraphics[width=1\linewidth]{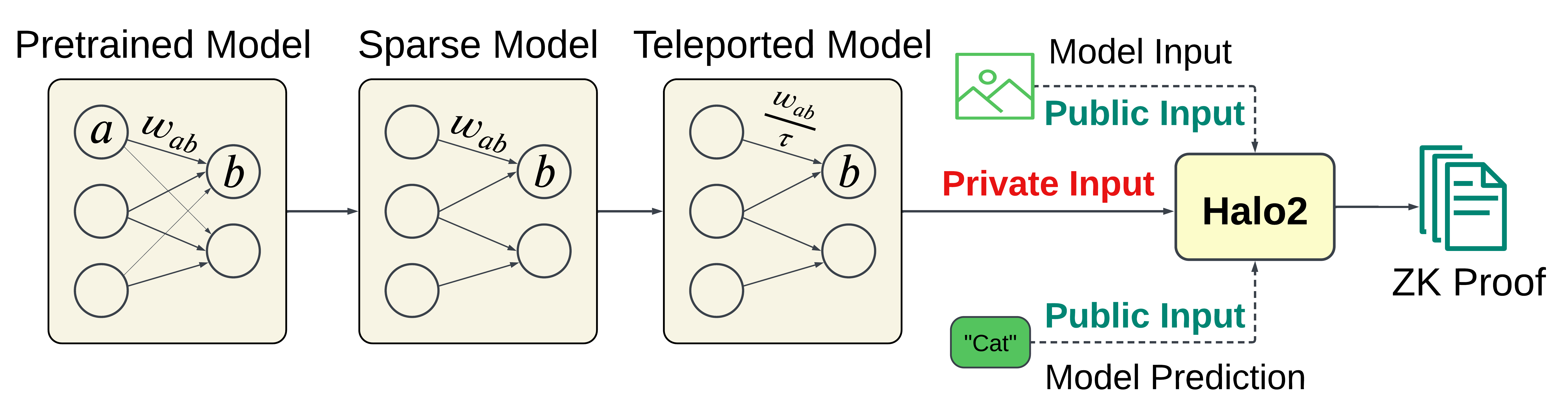}
    \caption{System overview: the neural network weight is fed as the private input to the Halo2 circuit. Input and output of the neural network is public input to the circuit.}
    \label{fig:overview}
\end{figure}

This section presents our methodology overview, outlining how each design goal from Section~\ref{sec:design_goal} is addressed. Section~\ref{sec:sparsification} describes our approach to reducing the number of circuit constraints by applying sparsification techniques to DNNs (design goal 1). We formalize the constraint-reduction process and its security analysis plus leverage two state-of-the-art post-pruning methods to illustrate its effectiveness in our experiments. In Section~\ref{sec:teleportation}, we use Neural Network Teleportation to constrain the range of activation functions, achieving design goal 2. 
The overview of the system is shown in Figure~\ref{fig:overview}, and the steps of the proposed \Name\ method for producing a ZK-friendly DNN are outlined in Algorithm~\ref{alg:methodology}.

\begin{algorithm}[t]
\caption{Overview of Methodology for Efficient Zero-Knowledge Proof Generation}
\label{alg:methodology}

\KwIn{Neural network weights $W$, input $X$, output $y$}
\KwOut{Valid zero-knowledge proof of computation represented by $\pi$ aligning with design goals ($G_1,G_2$)}

\vspace{0.5em}

\colonpoint{Sparsification (Addressing $G_1$)} \\
\vspace{0.2em}
Generate sparsified weights $W_{\text{sparse}}$ by solving:
\begin{equation*}
W_{\text{sparse}} \gets \arg\min_{\widetilde{W}} \sum_{i=1}^{L} \left\| f(x; W^{(i)}) - f(x; \widetilde{W}^{(i)}) \right\|^2
\end{equation*}
\hspace{2.5em} $\triangleright$ guided by pruning criteria (Eq.~\ref{eq:pruning_score}).

\vspace{1.0em}

\colonpoint{Teleportation (Addressing $G_2$)} \\
\vspace{0.2em}
Optimize CoB $\tau^\ast$:
\begin{equation*}
\tau^\ast \gets \arg\min_{\tau} \mathcal{L}(\tau) \quad \triangleright \text{where } \mathcal{L}(\tau) \text{ is defined in Eq.~\ref{eq:objective_function}};
\end{equation*}
\hspace{1.5em} (Solved by Eq.~\ref{eq:cge_general}, Eq.~\ref{eq:cge_estimate}, Eq.~\ref{eq:gradient_update}).\\
\vspace{0.2em}
Apply $\tau^\ast$ to $W_{\text{sparse}}$ (Eq.~\ref{eq:applying_weight_teleportaion}, Eq.~\ref{eq:scaled_activation}) to compute $W_{\text{sparse}}^t$.

\vspace{1.0em}

\colonpoint{ZK-SNARK Proof Generation} \\
\vspace{0.2em}
\begin{enumerate}
    \item Convert the model operations into Halo2 constraints (Eq.~\ref{eq:plonk_constraint}); 
    \item Eliminate constraints for indices where $W_{\text{sparse}}^t = 0$ to reduce circuit size. 
    \item Generate the zk-proof:
        \begin{equation*}
            \pi \gets \text{Halo2.Prove}(W_{\text{sparse}}^t, X, y)
        \end{equation*}
\end{enumerate}

\end{algorithm}

\section{Sparsification}
\label{sec:sparsification}
As noted in \ref{sec:design_goal}, we aim to optimize the required resource of the ZKP generation process by eliminating unnecessary constraints to achieve design goal 1. In the context of Halo2, we provide an example of one simple linear layer (fundamental layer in deep learning models) which is matrix-vector multiplication to show how sparsification of the model (matrix in this example) could reduce the number of required constraints without damaging security properties of ZK-SNARK. Specifically, we prove that omitting constraints corresponding to zero entries in the matrix (deep model weights) does not compromise the soundness of the proving system. We then present the two post-pruning methods employed in our experiments to complete the path towards reducing prover overhead by sparsification without degrading the accuracy of the DNN model.

\subsection{Preliminaries}

\point{Halo2 Proving System}
Halo2 employs an extended form of the PLONK arithmetization~\cite{gabizon2019plonk}, referred to as the “PLONKish”, which structures constraints in a polynomial format and introduces custom gates, lookup arguments, and permutation constraints. This framework is designed to operate over a matrix structure with cells, rows, and columns, where the rows correspond to elements in a multiplicative subgroup of a 
finite field $F_q$, 
typically sized as a power of two $(q=2^k)$.

Each circuit in Halo2 is represented by a matrix, where columns are categorized into:

\begin{itemize}
    \item \textbf{Fixed Columns (\( \text{Fixed} \)): } Columns with values fixed at circuit synthesis time, same across all proofs. Commitments to these columns are included in the verification key~\cite{zcash-halo2}.
    \item \textbf{Advice Columns (\( \text{Advice} \))}: Columns where the prover assigns values during proof generation. These values are private and known only to the prover.
    \item \textbf{Instance Columns (\text{Instance}): } Columns representing public inputs to the circuit, provided by the verifier and fixed at proof time. These columns facilitate the verification of statements involving known values.
\end{itemize}

In addition to these columns, polynomial commitments are utilized to secure the integrity of the data within these columns. A commitment scheme, such as KZG commitments~\cite{kate2010constant}, is employed to bind the polynomials that represent each column. 

During proof creation, the prover sets up a matrix with \textit{advice, instance, and fixed columns} and populates each cell with field elements, denoted \( A_{i,j} \) for advice or \( F_{i,j} \) for fixed values in j-th row and i-th column. To commit to these values, the prover constructs Lagrange polynomials of degree \( n-1 \) for each column. For advice and fixed columns, the Lagrange polynomials \( a_i(X) \) and \( q_i(X) \) are interpolated over the evaluation domain of size \( n \), where \( \omega \) is the \( n \)-th primitive root of unity:
\[
   a_i(X) \text{ such that } a_i(\omega^j) = A_{i,j},
\]
\[
   q_i(X) \text{ such that } q_i(\omega^j) = F_{i,j}.
\]
Commitments to these polynomials are then made as:
\[
   A = [\text{Commit}(a_0(X)), \ldots, \text{Commit}(a_i(X))],
\]
\begin{equation}
\label{eq:verificationkey}
    Q = [\text{Commit}(q_0(X)), \ldots, \text{Commit}(q_i(X))],
\end{equation}
   
where \( Q \) is established during key generation, while \( A \) is produced by the prover and sent to the verifier.

Halo2 enforces three primary types of constraints:

\begin{itemize}
    \item \textbf{Custom Gates:} These define polynomial constraints within rows, enabling expressions such as multiplication and addition to be enforced across specific rows. 
    The generic form of constraints in PLONK (and similarly used in Halo2) is formulated as:

    \begin{align}
        \label{eq:plonk_constraint}
        & a_i(X) \cdot q_A(X) 
        + b_{i'}(X) \cdot q_B(X) \nonumber \\
        & + a_i(X) \cdot b_{i'}(X) \cdot q_M(X) 
        + c_{i''}(X) \cdot q_C(X) = 0,
    \end{align}
where \( a_i(X) \), \( b_{i'}(X) \), and \( c_{i''}(X) \) represent the polynomial assignments in advice columns \( i,i',i'' \) of the arithmetization matrix, and \( q_A(X) \), \( q_B(X) \), \( q_M(X) \), and \( q_C(X) \) are the corresponding polynomials which are stored in fixed columns.

    \item \textbf{Permutation Arguments:} These allow cell values to match across different locations in the matrix, using randomized polynomial constraints for multi-set equality checks. Permutation arguments enable values to be "copied" within the circuit.
   
    \item \textbf{Lookup Arguments:} Lookup arguments constrain a \( k \)-tuple of cells \( (A_{1,j}, \ldots, A_{k,j}) \) in the same row \( j \) such that for a disjoint set of \( k \) columns, these cells match the values in some other row \( j' \). We can enforce the following constraint:

\[
(A_{1,j}, \ldots, d_{k,j}) = (A_{1',j'}, \ldots, A_{k',j'}),
\]

ensuring \( (A_{1,j}, \ldots, A_{k,j}) \) lies within the lookup table defined by those \( k \) columns~\cite{zcash-halo2,kang2022scaling}.
\end{itemize}


The formal ZK-SNARK properties guaranteed by Halo2 and the commitment schema are detailed in Appendix~\ref{sec:appendix_all_properties}.

\subsection{Matrix-Vector Multiplication in Halo2}
In this section, we formulate circuit constraints for sparse model weights, aiming to reduce the overall number of required constraints. Without loss of generality, we focus on a fully connected (linear) layer in neural networks. In this setting, the model parameters (weights of the linear layer) are represented by a matrix \( W \in \mathbb{R}^{d_1 \times d_2} \), and the input to the layer (the output from the previous layer) is represented by a vector \( I \in \mathbb{R}^{d_2} \). The output of the layer is denoted by vector \( O \in \mathbb{R}^{d_1} \), calculated as:

\[
O = W \cdot I
\]
Each component \( o_i \) of the output vector \( o \) is computed as:
\begin{equation}
\label{eq:matrix_vector_mul}
o_i = \sum_{j=1}^{d_2} W_{i,j} \cdot I_j
\end{equation}

By leveraging the sparsity of \( W \), we can reduce the number of constraints in the circuit, 
thereby decreasing the computations required for verification. In the following we explore how it would be possible for Halo2 to do that efficiently without compromising the soundness of Halo2 mentioned in Equation~[\ref{eq:zk_soundness}].

\point{Circuit Representation}
To construct a ZKP over the matrix-vector multiplication of Equation~\eqref{eq:matrix_vector_mul}, we should describe the equation using Halo2 circuit constraints using the generic constraints form presented in equation~\ref{eq:plonk_constraint}.

\begin{figure}
    \centering
    \includegraphics[width=1\linewidth]{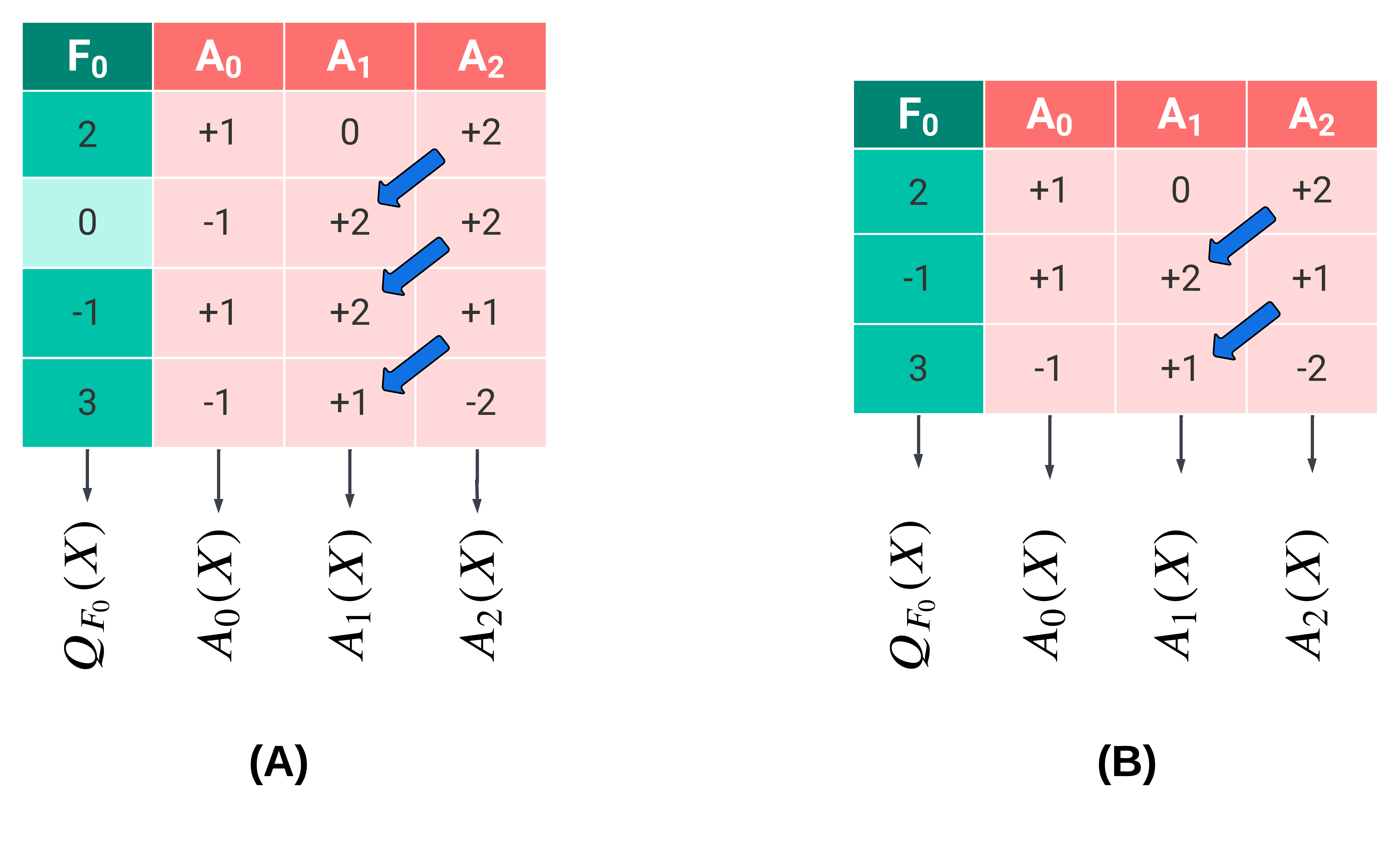}
    \caption{(A) represent a Halo2 circuit without removing sparse weights constraints while (B) represents it after removing zero-weight entries.}
    \label{fig:halo2_sparse_table}
\end{figure}

To provide proof generation on the Linear Layer,  as shown in figure~\ref{fig:halo2_sparse_table}, we put the model weights $W_{i,j}$ in a fixed column of the Halo2 table named $F_0$ and input of the current layer ($I_j$) to an advice column named $A_0$. To provide the outputs we consider two advice columns $A_1$ and $A_2$ such that for all rows $i \in d_2$ the following equation should be kept for table cells:

\begin{equation}
    \label{eq:table_constraints_linearlayer}
    F_{0i} * A_{0i} + A_{1i} = A_{2i}
\end{equation}

Note that these constraints followed the form of plonkish constrained previously mentioned in~\ref{eq:plonk_constraint}. By interpolation of the polynomial function, these constraints will be enforced by the following polynomial equation which should be validated by the prover:
\begin{equation}
    \label{eq:polynomial_constraints_linearlayer}
    A_{0}(X) \cdot Q_{F_0} (X) + A_{1}(X) - A_{2}(X) = 0
\end{equation}

As illustrated in Figure~\ref{fig:halo2_sparse_table}, copy constraints are applied between advice columns  $A_1$  and  $A_2$ . These constraints ensure that the accumulated value in row  $i$  of  $A_2$  is correctly propagated to row  $i+1$  in  $A_1$ , preserving value integrity across the table. Furthermore, commitments to the fixed column,  $\text{Commit}(Q_{F_0}(X))$ , as well as the advice columns,  $\text{Commit}(A_1(X))$  and  $\text{Commit}(A_2(X))$ , are included in the verification key and proof, respectively.

So far, we have shown how a dense linear neural network layer is represented using Halo2 constraints. Next, we explore reducing constraints—and rows in the table—when model weights are sparse, enhancing efficiency without compromising Halo2's soundness.

\point{Handling Zero Entries}
Based on equation~\ref{eq:matrix_vector_mul}, when \( W_{i,j} = 0 \), the term \( W_{i,j} I_j = 0 \) does not contribute to the corresponding output $o_{i}$ while preserving the correctness of the equation. Therefore, we can eliminate constraints involving zero entries so as not to include constraints for terms where \( W_{i,j} = 0 \). In particular, we could eliminate the row $t$ of the table in equation~\ref{eq:table_constraints_linearlayer} where $F_{0t}$ is equal to zero. 
This reduction decreases the number of constraints and computations required during proof generation.
In the following, we formalize this constraint reduction.

\begin{theorem}
In a matrix-vector multiplication circuit in Halo2, where the matrix \( W \) is stored in fixed columns, eliminating constraints corresponding to zero entries \( W_{i,j} = 0 \) does not compromise the soundness of the proving system.
\label{theorem:sparsity}
\end{theorem}

The proof of Theorem~\ref{theorem:sparsity} can be found in Appendix~\ref{sec:appendix_formal_proof}.

\subsection{DNN Sparsification}

The goal of model sparsification, given a dense model with parameters \( W \in \mathbb{R}^d \), is to identify a subset of weights that can be zeroed out while still maintaining the model's overall performance. We focus on \textit{post-training sparsification methods}, meaning the model has already been trained, and pruning is applied afterwards. This allows us to preserve the original training process while optimizing the model for ZKP generation. 
Model pruning is performed offline before zkSNARK proving, independently of the ZK circuit inputs. By reusing a single pruned model for multiple proofs, per-run CPU benchmarks are not a bottleneck, so we exclude their costs.
We leverage two state-of-the-art pruning methods, \textbf{RD\_PRUNE}~\cite{xu2023efficient} and \textbf{CAP}~\cite{kuznedelev2024cap}. These methods offer efficient execution even on CPUs, enabling the sparsification of large models in a matter of seconds, without significant loss of model accuracy.


Pruning techniques typically compute an importance score for each weight and then remove the least important weights based on this score. The pruning objective can be summarized as:

\begin{equation}
    \label{eq:sparsification_optimization}
    \min_{\mathbf{W}} \sum_{i=1}^{L} \mathcal{L}(f(\mathbf{W}^{(i)})) + \lambda \|\mathbf{W}^{(i)}\|_0
\end{equation}
where  $\mathbf{W}^{(i)}$  represents the weights of layer  $i$,  $\mathcal{L}$  is the loss function, and  $\|\mathbf{W}^{(i)}\|_0$  imposes a sparsity constraint by penalizing the number of non-zero weights. The loss function  $\mathcal{L}(f(\mathbf{W}^{(i)}))$  measures the error or deviation of the model's predictions  $f(\mathbf{W}^{(i)})$  from the true outputs, quantifying the model's performance while excluding the pruned weights. By minimizing this loss, the model seeks to retain performance even after pruning. 
Several metrics have been explored to measure weight importance in prior works. Weight magnitude is one of the most commonly used criteria~\cite{gale2019state,zhu2017prune} where in each iteration, weights with the lowest magnitude are pruned.
Recent approaches have utilized second-order information about the loss to determine the importance of each weight, resulting in better preservation of the model's performance especially in transformer-based architectures~\cite{kuznedelev2024cap,dong2017learning,wang2019eigendamage,yu2022hessian}.

Building on these ideas, RD\_PRUNE and CAP introduce further refinements by treating the model as a sequence of blocks, thereby making the pruning process layer-wise and computationally efficient. CAP, in particular, uses an approximation of second-order information as its pruning metric, allowing it to maintain model accuracy more effectively. In contrast, RD\_PRUNE leverages output distortion as its pruning metric, focusing on the reconstruction of the network's output behaviour rather than relying on the weight magnitudes or approximated second-order information.

The first method introduces the concept of \textit{distortion}, which quantifies the impact of pruning on the model's output. The pruning problem is framed as an optimization problem, where the objective is to minimize the overall distortion across layers. This can be expressed as:

\begin{equation}
    \label{eq:detailed_saprse_optimization}
    \min \sum_{i=1}^{L} \left\| f(x; W^{(i)}) - f(x; \widetilde{W}^{(i)}) \right\|^2 
\quad \text{s.t.} \quad
\sum_{i=1}^{L} \frac{\|\widetilde{W}^{(i)}\|_0}{\|W^{(i)}\|_0} \leq R
\end{equation}
where $\widetilde{W}$ and $\left\| f(x; W^{(i)}) - f(x; \widetilde{W}^{(i)}) \right\|^2$ represent the pruned model weights  and the distortion at layer \( i \) after pruning respectively.  $R$ is the sparsity ratio of the entire network. RD\_PRUNE uses \textit{dynamic programming} to solve this optimization problem, determining the optimal pruning for each layer. The method follows a \textit{one-shot pruning} setting to fine-tune the pruned model by small calibration samples $\mathcal{D}_{\text{calib}}$ (e.g., 1024 samples) to retain the model performance.

\point{CAP (Correlation-Aware Pruning)}
The second method takes a more sophisticated pruning criterion based on \textit{weight correlations}. This method uses a \textit{block-wise Hessian update}, which makes it computationally efficient, as it approximates the Hessian matrix in a block-diagonal form, significantly reducing the computational complexity of the pruning process. CAP can prune models in a \textit{zero-shot setting}, meaning no additional training data is required after pruning, further enhancing its efficiency, especially in large models like ViT. In particular, CAP minimizes the following objective:

\[
\min_{W} \frac{1}{2} \sum_{i=1}^{L} \left( W_i^T H_i^{-1} W_i \right)
\]
where \( \mathbf{H}_i \) is the Hessian matrix for block \( i \). By using a block-diagonal approximation of the Hessian, CAP iteratively updates the pruning scores for each block, ensuring fast computation while maintaining high accuracy. Within each block, CAP calculates the importance of the \( i \)-th weight as:

\begin{equation}
    \label{eq:pruning_score}
    \rho_i = \frac{w_i^2}{2 \left[ H_L^{-1}(w, \mathcal{D}_{\text{calib}}) \right]_{ii}}
\end{equation}

Weights with the smallest importance scores \( \rho_i \) are pruned iteratively. To mitigate the impact of pruning, the optimal update for the remaining weights in the block is computed using:

\[
\delta w = - \frac{w_i}{\left[ H_L^{-1}(w) \right]_{ii}} H_L^{-1}(w,\mathcal{D}_{\text{calib}}) e_i
\]

This iterative pruning process repeats across all blocks, progressively pruning the least important weights. By using this block-wise approach, CAP effectively balances computational efficiency with model accuracy, making it well-suited for large-scale model pruning without requiring retraining epochs or additional data.

\section{Teleportation}
\label{sec:teleportation}

Deep neural networks (DNNs), especially architectures like modern ViTs that utilize Layer Normalization, often produce outlier values in the inputs to activation functions due to the normalization process~\cite{ding2022towards,liu2023noisyquant,ashkboos2024quarot}. In most of the recently utilized ZK-proving systems for DNNs, constraints on activation functions are typically represented using lookup tables~\cite{ezkl,sun2024zkllm,zcash-halo2,kang2022scaling}. Outlier values broaden the input range of activation functions, necessitating larger lookup tables. This expansion increases proving time, memory usage, and proof size as shown in experiment section. Our methodology aims to mitigate these outlier values and optimize the activation input range by leveraging neural teleportation, thus enhancing the efficiency of the verification process. Neural teleportation is a mathematical framework that transforms network parameters while preserving the network's function, thereby reducing the range of activation inputs without affecting the network's outputs.

Neural teleportation assigns positive scaling factors, or \textbf{Change of Basis (CoB)} scalars $\tau_j^{(i)} > 0$, to each \emph{layer} $i$ and \emph{neuron} $j$ in that layer, where $i \in \{1, \dots, L\}$ and $L$ is the total number of layers. The weights $w_{j,k}^{(i)}$ connecting neuron $j$ in layer $i$ to neuron $k$ in layer $i+1$ are transformed as~\cite{armenta2023neural,zhao2022symmetry}:

\begin{equation}
\label{eq:applying_weight_teleportaion}
v_{j,k}^{(i)} = \left( \frac{\tau_k^{(i+1)}}{\tau_j^{(i)}} \right) w_{j,k}^{(i)}
\end{equation}
The activation functions $f_j^{(i)}$ at each neuron are adjusted to:

\begin{equation}
\label{eq:scaled_activation}
g_j^{(i)}(x) = \tau_j^{(i)} f_j^{(i)}\left( \frac{x}{\tau_j^{(i)}} \right)
\end{equation}

An activation function $f$ is \textbf{positive scale-invariant} if:

\begin{equation}
f(c x) = c f(x), \quad \forall x \in \mathbb{R}, \quad \forall c > 0
\end{equation}
This property ensures that scaling the input by a positive factor $c$ scales the output by the same factor, which is critical for maintaining consistency in neural transformations.
A common example is the ReLU function, $f(x) = \max(0, x)$, widely used due to its simplicity and effectiveness in DNNs. For such functions $f$, any transformed activation simplifies back to the original, preserving the output of the network:

\begin{equation}
\label{eq:scale_invariance}
g_j^{(i)}(x) = \tau_j^{(i)} f_j^{(i)}\left( \frac{x}{\tau_j^{(i)}} \right) = f_j^{(i)}(x)
\end{equation}
Ensuring the network's output stays the same after transformation. Our goal is to optimize the CoB scalars $\{\tau_j^{(i)}\}$ to minimize the range of scaled pre-activation inputs $\frac{z_j^{(i)}}{\tau_j^{(i)}}$ across all activation functions in different layers. Specifically, we aim to minimize the difference between the maximum and minimum scaled pre-activation inputs within each layer (or block) $i$:

\begin{equation}
\label{eq:range_optimization}
\min_{{ {\tau > 0}}} \quad \sum_{i=1}^{L} \left( \max_{j} \left( \frac{z_{j}^{(i)}}{\tau_j^{(i)}} \right) - \min_{j} \left( \frac{z_{j}^{(i)}}{\tau_j^{(i)}} \right) \right)
\end{equation}
which $z_j^{(i)}$ representing input of $j$-th neuron in the layer $i$.

The CoB constraints, derived from Theorem 2.1 in previous work~\cite{armenta2023neural}, are applied as follows:
\begin{enumerate}
\item \textbf{Input, Output, and Bias Layers}: Cob scalars $\{\tau_j^{(i)}\}$ are assigned if layer $i$ corresponds to input, output, or bias layers.
\item \textbf{Residual Connections}: For layers $i$ and $j$ connected by residual paths, $\{\tau_j^{(i)}\}$ are assigned.
\item \textbf{Convolutional Layers}: Neurons within the same feature map share the same $\tau_i$.
\item \textbf{Batch Normalization Layers}: CoB scalars are applied to parameters $\gamma$ and $\beta$, but not to the running mean and variance.
\end{enumerate}

To achieve a reduced range of activation function inputs, We define the objective function $l(\tau)$ as:

\begin{equation}
\label{eq:objective_function}
l(\tau) = \sum_{i=1}^{L} \left( \max_j \left( \frac{z_j^{(i)}}{\tau_j^{(i)}} \right) - \min_j \left( \frac{z_j^{(i)}}{\tau_j^{(i)}} \right) \right)
\end{equation}
Then our optimization problem summarized as:

\begin{equation}
   \min_{\tau > 0} \quad l(\tau) 
\end{equation}

Optimizing the CoB scalars reduces the range of activation inputs within each layer, thereby minimizing the impact of outlier values and reducing the ZK lookup table overhead.
Despite the significant reduction in parameters ${\tau_j^{(i)}}$ to optimize compared to model weights (e.g., 10K vs. 5 million in Tiny ViT), the optimization process in Equation~\eqref{eq:range_optimization} presents two key challenges:

\begin{itemize}
    \item \textbf{Non-Differentiable Objective}: The $\max$ and $\min$ functions introduce discontinuities in the gradient with respect to $\tau$. As a result, traditional gradient-based optimization methods are ineffective at such points, necessitating alternative approaches.
    \item \textbf{Interdependent Scaling Factors}: The interconnected nature of neural networks means that adjusting one scaling factor $\tau_j^{(i)}$ influences other activations. This interdependence causes the indices $j$ for $\max_j$ and $\min_j$ to vary dynamically, depending on $\tau$, the input data, and the network weights. Such variations further complicate optimization, as the maxima and minima are not fixed during the process.
\end{itemize}

To overcome these challenges, we use Coordinate Gradient Estimation (CGE)~\cite{chen2023deepzero}, a zero-order optimization method. Zero-order methods approximate gradients using function evaluations, avoiding explicit gradient computations. The general form of CGE is:

\begin{equation}
\label{eq:cge_general}
\hat{\nabla}_{\tau} \ell(\tau) = \sum_{i=1}^{d} \left[ \frac{\ell(\tau + \mu \mathbf{e}_i) - \ell(\tau)}{\mu} \mathbf{e}_i \right]
\end{equation}
where $\mu > 0$ is a small perturbation scalar. $\mathbf{e}_i$ is the i-th standard basis vector in $\mathbb{R}^{d}$. 
Applying this to our optimization problem, we approximate the gradient with respect to each $\tau_j^{(i)}$ as:

\begin{equation}
\label{eq:cge_estimate}
\hat{g}_j = \frac{\ell(\tau + \mu \mathbf{e}_j) - \ell(\tau)}{\mu}
\end{equation}


Using this estimated gradient, we perform gradient descent updates on the CoB scalars:

\begin{equation}
\label{eq:gradient_update}
\tau_j^{(i)} \leftarrow \tau_j^{(i)} - \eta \hat{g}_j
\end{equation}
where $\eta > 0$ is the learning rate, and $\hat{g}_j$ is the $j$-th component of $\hat{\nabla}_{\tau} \ell(\tau)$.
Since $\tau_j^{(i)} > 0$, after each update, We project $\tau_j^{(i)}$ onto the positive orthant.

\subsection{Advantages of Zero-Order Methods}
Zero-order methods offer several advantages that make them particularly well-suited for the defined optimization task: 
\begin{itemize}
    \item \textbf{No Need for Backpropagation}: Zero-order methods do not require gradient computations via backpropagation, which can be memory and computationally intensive for large models.
    \item \textbf{Handling Non-Differentiable Functions}: Zero-order methods are suitable for optimizing non-differentiable functions, making them appropriate for our problem.
    \item \textbf{Parallelization}: The gradient estimation procedure can be parallelized since each coordinate perturbation is independent. This enables teleportation optimization to be completed in a reasonable time frame compared to the proof generation, as explained in the experimental section.
    
\end{itemize}

\subsection{Extension to Non-Scale-Invariant Activation}
In Equation~\eqref{eq:scale_invariance}, we assumed that the activation functions used in the DNN are positive scale-invariant. However, many modern DNNs employ activation functions that are not strictly scale-invariant, such as the Gaussian Error Linear Unit (GELU). The GELU activation function is defined as:

\begin{equation}
\label{eq:gelu_definition}
\mathrm{GELU}(x) = x \cdot \Phi(x) = x \cdot \frac{1}{2} \left( 1 + \operatorname{erf}\left( \frac{x}{\sqrt{2}} \right) \right)
\end{equation}
where $\Phi(x)$ is the cumulative distribution function of the standard normal distribution, and \(\operatorname{erf}\) is the error function.

Assuming that such non-scale-invariant activation functions are scale-invariant can lead to changes in the function that the neural network represents, thus altering its function. To address this challenge, we consider two important aspects:

	1.	\textbf{Approximate Scale-Invariance at Extremes}: Non-scale-invariant activation functions like GELU become approximately scale-invariant for large positive or negative inputs. Specifically, as $x \to \infty$ or $x \to -\infty$, the GELU function behaves similarly to the ReLU or linear functions, respectively, which are scale-invariant.
 
	2.	\textbf{Minimizing Approximation Error}: To reduce the effect of the approximation error introduced by assuming scale invariance, we add a reconstruction term to the optimization problem in Equation~\eqref{eq:range_optimization}. This term penalizes the difference between the outputs of the original model and the teleported model, ensuring that the teleported model remains close in function space to the original.

Incorporating these considerations, we update our optimization problem to include the reconstruction term:

\begin{equation}
\label{eq:updated_optimization}
\min_{\tau > 0} \quad \sum_{i=1}^{L} \left( \max_j \left( \frac{z_j^{(i)}}{\tau_j^{(i)}} \right) - \min_j \left( \frac{z_j^{(i)}}{\tau_j^{(i)}} \right) \right) + \lambda \left| f(\boldsymbol{\theta}, \mathbf{1}) - f(\boldsymbol{\theta}, \tau) \right|^2
\end{equation}

\begin{itemize}
    \item $ f(\boldsymbol{\theta}, \mathbf{1}) $ represents the original neural network function evaluated with scaling factors  $\tau = \mathbf{1}$  (i.e., no scaling applied).
    \item $f(\boldsymbol{\theta}, \tau)$ represents the teleported neural network function evaluated with the optimized scaling factors  $\tau$.
    \item $\lambda > 0$ is a regularization hyperparameter that balances the trade-off between minimizing the activation input range and preserving the original network function.
\end{itemize}

The reconstruction term 
$\left| f(\boldsymbol{\theta}, \mathbf{1}) - f(\boldsymbol{\theta}, \tau) \right|^2$
quantifies the difference between the outputs of the original and teleported models, thereby controlling the approximation error due to the non-scale-invariant activation functions.

The algorithm outlined in Algorithm~\ref{alg:CGE} guides the optimization process towards convergence on CoB scalers, effectively reducing the activation input ranges. This reduces the impact of large lookup tables, enhancing resource efficiency as shown in the experiments.

\begin{algorithm}[t]
\caption{Optimization teleportation parameters via CGE}
\label{alg:CGE}
\KwIn{Initial scaling factors $\tau = \{\tau_j^{(i)}\}$, learning rate $\eta$, perturbation size $\mu$, max iterations $N$}
\KwOut{Optimized scaling factors $\tau^\ast$}
\For{$n = 1$ \KwTo $N$}{
    Evaluate objective function $l(\tau)$ using Eq.~\eqref{eq:updated_optimization}.\;
    \ForEach{coordinate $\tau_j^{(i)}$ \textbf{(parallelizable)}}{
        Compute gradient estimate:
        \begin{equation*}
            \text{Evaluate} \ l(\tau + \mu \mathbf{e}_j)
        \end{equation*}
        \begin{equation*}
            \hat{g}_j = \frac{l(\tau + \mu \mathbf{e}_j) - l(\tau)}{\mu}
        \end{equation*}
    }
    Update scaling factors:
    \begin{equation*}
    \tau_j^{(i)} \leftarrow \tau_j^{(i)} - \eta \hat{g}_j
    \end{equation*}
    Apply constraints:
    \begin{equation*}
    \tau_j^{(i)} \leftarrow \max(\tau_j^{(i)}, \epsilon)
    \end{equation*}
    Enforce CoB constraints (see Section~3).\;
    \If{Convergence criteria met (\emph{e.g.}, $\| \hat{\nabla}_{\tau} l(\tau) \| < \delta$)}{
        \textbf{break}\;
    }
}
\Return Optimized scaling factors $\tau^\ast = \tau$\;
\end{algorithm}

\section{Experiments}
\label{sec:experiments}

\subsection{Implementation}

To demonstrate the efficiency of \Name\ in reducing resource consumption for ZKP generation, we conducted experiments on both dense and sparse versions of DNN models. Our focus is on assessing the resource usage of the prover and verifier under each configuration, specifically observing how the proposed post-processing of \Name\ impacts computation time and memory demands. To measure that, we employed the EZKL toolkit~\cite{ezkl}, a state-of-the-art tool for generating ZK-SNARK proofs for DNN models, which is based on Halo2. This toolkit utilizes fixed-point quantization, akin to previous methods exploring ZK-SNARKs of DNNs~\cite{kang2022scaling}.

\subsection{Relevance of Selected state-of-the-art}
In order to benchmark \Name\ against previous State-of-the-art works considering verifiable inference by ZK-SNARK, we selected the zero-knowledge proving system outlined in~\cite{kang2022scaling} as a reference. This system, based on Halo2, supports a more diverse range of modern DNN architectures rather than previous works, which aligns with our goal of evaluating scalability across various architectures. We chose not to include CNN-specific methods such as~\cite{liu2021zkcnn, lee2024vcnn} in our evaluation, as they do not generalize to the types of DNNs considered in our study. Since our experiments were conducted in a CPU-only environment, we did not include GPU-dependent frameworks, such as the system proposed in~\cite{sun2024zkllm}, as they would not provide a directly comparable assessment of CPU performance. Additionally, as discussed in Section~\ref{sec:introduction}, \Name\ is designed to be complementary to previously proposed ZKP systems that are tailored ZKP for specific architectures. \Name\ can be applied on top of existing specialized frameworks to achieve even more optimized performance.

\subsection{Results}
For consistency, all experiments were conducted on a virtual machine running Linux (kernel 4.18.0), equipped with 32 virtual CPU cores, 1 TB of memory, and 2 GB of swap space.
The fixed-point quantization scale was set at $2^{12}$ to balance performance and memory usage.
Increasing this scale would enhance model accuracy but also increase memory usage during proof generation due to the increasing the required number of field values and larger lookup tables required for non-linear functions in the proving system~\cite{kang2022scaling,ezkl}.

To evaluate the effectiveness of \Name, we conducted experiments on three distinct setups: MobileNetV1~\cite{howard2017mobilenets} on the CIFAR-10 dataset~\cite{krizhevsky2009learning} and ResNet-20~\cite{he2016deep} on the CIFAR-100 dataset, as detailed in Table~\ref{tab:framework_comparison_combined}, and the ViT model~\cite{dosovitskiy2020image} on the large-scale ImageNet dataset~\cite{deng2009imagenet}, as presented in Table~\ref{tab:main_performance}. This selection includes varied architectures and datasets, providing a comprehensive evaluation of our method's adaptability and efficiency.

\begin{table*}[tbh]
\centering
\adjustbox{max width=\textwidth}{
\begin{tabular}{@{}llccccc@{}}
\toprule
Model & Dataset & Framework & \begin{tabular}[c]{@{}c@{}}Avg Memory Usage \\ (GB) ± Std\end{tabular} & \begin{tabular}[c]{@{}c@{}}Avg Proving Time \\ (s) ± Std\end{tabular} & \begin{tabular}[c]{@{}c@{}}Verification Time \\ (ms)\end{tabular} & \begin{tabular}[c]{@{}c@{}}Proof Size \\ (KB)\end{tabular} \\ \midrule

\multirow{4}{*}{\centering MobileNetv1} & \multirow{4}{*}{\centering CIFAR-10} 
& ZKML (tensorflow) & 148.3 ± 10.2 & 1439 & 23.8 & 106 \\
&  & EZKL (pytorch) & 139.0 ± 14.2 & 779 & 3.5 & 100 \\
&  & \textbf{\Name} (sparsity=50\%) & \textbf{56.7 ± 3.0}  & \textbf{358} & \textbf{2.7} & \textbf{70} \\ \cmidrule(l){3-7}
&  & \textbf{Reduction} (\%) & \textbf{59.2\%} & \textbf{54.0\%} & \textbf{22.9\%} & \textbf{30.0\%} \\ \midrule[1.5pt]

\multirow{4}{*}{\centering Resnet-20} & \multirow{4}{*}{\centering CIFAR-100} 
& ZKML (tensorflow) & 128.0 ± 7.5 & 1055 & 20.1 & 89 \\
&  & EZKL (pytorch) & 120.2 ± 5.5 & 564 & 3.1 & 85 \\
&  & \textbf{\Name} (sparsity=50\%) & \textbf{39.8 ± 3.1} & \textbf{307} & \textbf{2.5} & \textbf{59} \\ \cmidrule(l){3-7}
&  & \textbf{Reduction} (\%) & \textbf{66.8\%} & \textbf{45.6\%} & \textbf{19.4\%} & \textbf{30.6\%} \\ \bottomrule

\end{tabular}
}
\caption{Comparison of various frameworks across models and datasets, with reduction percentage achieved by \Name\ compared to prior work. The standard deviation is computed by repeating the experiment five times.}
\label{tab:framework_comparison_combined}
\end{table*}

As shown in Table~\ref{tab:framework_comparison_combined}, we present the proving time, peak memory usage of the prover, and proof size for both MobileNetv1 on CIFAR-10 and ResNet-20 on CIFAR-100, showcasing the effectiveness of the proposed method. With a $50\%$ sparsity ratio and applying the optimized teleportation, our approach balances model performance (detailed in Table~\ref{tab:acc}) and the reduction of resource consumption. Importantly, the sparsity consideration is unique to \Name\, as prior works like~\cite{kang2022scaling} are not designed for sparse models and therefore do not benefit from the associated resource reductions.

For MobileNetv1, \Name\ reduces average memory usage by 59.2\% and proving time by 54.0\% compared to EZKL without the proposed post-processing. The proof size and verification time are reduced by 30.0\% and 22.9\%, demonstrating a comprehensive improvement in computational and bandwidth usage of the verifier. 
For ResNet-20, memory usage decreases by 66.8\%, proving time by 45.6\%, proof size by 30.6\%, and verification time by 19.4\%.
The improvements are due to sparsification (Section~\ref{sec:sparsification}) and teleportation (Section~\ref{sec:teleportation}), with their individual effects analyzed in the ablation study in Table~\ref{tab:ablation}.

In contrast to prior work focusing on small datasets and simplified model architectures, our experiments utilize the large-scale ImageNet dataset~\cite{deng2009imagenet} and the ViT transformer-based~\cite{dosovitskiy2020image} architecture for a thorough evaluation. Given that ViT's computational demands are substantial—with approximately 31 times more FLOPs compared to Resnet-20 — we employed circuit splitting technique~\cite{ezkl} (detailed in~\ref{sec:appendix_splitting}) to make proof generation practical, reducing memory overhead for the prover while maintaining model integrity.
In terms of splitting the model, we split the model into $M=24$ parts for the ViT model.
This decision was motivated by two key factors: Firstly, splitting each layer into two components (self-attention and MLP) is advantageous as the intermediate model output is relatively smaller compared to other splitting options, thereby reducing the overhead of commitment to intermediate results. Secondly, the primary factor influencing proving resource usage is the logarithm of the number of rows in the halo2 proving table~\cite{south2024verifiable}. 
By splitting the model such that it features a roughly similar number of constraints across all parts, ensuring each component has an identical logarithmic number of rows. Consequently, memory usage is distributed more uniformly across circuit parts, thereby reducing the maximum memory size required during proof generation. Moreover, if proof generation exhibits similar time across circuit components, parallelization becomes more efficient, and the time required is equivalent to generating only one proof. Although the experimental results presented do not include parallelization, it's worth noting that parallel proof generation could be achieved because each segment operates with distinct inputs and independently.

\begin{table}[tbh]
\centering
\adjustbox{max width=\columnwidth}{
\begin{tabular}{@{}lcccc@{}}
\toprule
Framework & \begin{tabular}[c]{@{}c@{}}Mem. Usage \\ (GB)\end{tabular} & \begin{tabular}[c]{@{}c@{}}Proving \\ (s)\end{tabular} & \begin{tabular}[c]{@{}c@{}}Verification \\ (ms)\end{tabular} & \begin{tabular}[c]{@{}c@{}}Proof Size \\ (KB)\end{tabular} \\ \midrule
EZKL (pytorch) & 650 & 1087 & 4.5 & 1134 \\
\textbf{\Name} (50\%) & \textbf{357} & \textbf{869} & \textbf{4.1} & \textbf{1076} \\\midrule
\textbf{Reduction} (\%) & \textbf{45.08\%} & \textbf{20.05\%} & \textbf{8.89\%} & \textbf{5.12\%} \\ \bottomrule
\end{tabular}
}
\caption{Comparison of frameworks for Tiny-ViT on ImageNet. Each column represents the average across $M=24$ model splits, the reduction is calculated relative to EZKL.}
\label{tab:main_performance}
\vspace{-15pt}
\end{table}

Table~\ref{tab:main_performance} highlights the effectiveness of \Name\ in reducing proving time and memory usage on splitted circuits of the ViT model. Each column shows the corresponding metric averaged across $M$ parts of the model. 
\Name\ effectively reduces memory usage in transformer architectures too, achieving a notable 45.08\% reduction even when applied to model splitting. However, proof generation for ViT demands ~6 times more memory than MobileNet-v1, likely due to ViT's computational intensity, with approximately 5.5 GFLOPs, making it unsuitable for resource-constrained devices. On the positive side, verification time remains efficient, increasing only ~1.44 times compared to MobileNet-v1, highlighting the scalability of the verification process despite the proving overhead. 

Existing approaches \cite{weng2023pvcnn,fan2023validating} are not directly applicable with Transformers and modern CNNs like MobileNet. \cite{weng2023pvcnn} performed less favorably compared to ZKML \cite{kang2022scaling}, as shown in Table 2 of the ZKML paper, while our results compare favorably. To have a comparison with TeleSparse, we tested \cite{weng2023pvcnn} on LeNet-5 using similar hardware. Compared to \cite{weng2023pvcnn} (127.2s) and \cite{fan2023validating} (11.6s), TeleSparse achieved 7.4s ± 0.07, demonstrating superior efficiency.

The quantization applied to weights and activation values within the ZK proving system, combined with the sparsification techniques we employed, would decrease model accuracy within the circuit. To evaluate that, Section~\ref{sec:accuracy} examines the trade-offs in accuracy. Additionally, Section~\ref{sec:ablation} analyzes the individual contributions of each component of \Name\ to the resource reductions presented in the tables. Notably, the resource usage for teleportation, included in Tables~\ref{tab:framework_comparison_combined} and~\ref{tab:main_performance}, remains minimal. Even for ViT, teleportation requires less than 2\% of the memory needed for proof generation, and the total teleportation time for all MLP parts (to which teleportation exclusively applies) is ~50 seconds, constituting less than 0.1\% of the overall proof generation time for the model.

\subsection{Accuracy}
\label{sec:accuracy}
To assess the impact of quantization and post-processing introduced by \Name\ on model performance over the ZKP system, we evaluate both the original floating-point model accuracy and its quantized model (and post-processed) accuracy provided by the ZKP framework in Table~\ref{tab:acc}. 

By selecting to be $2^{12}$, the results show that the model retains strong performance even with pruning. The difference between the full-precision model and the ZK model is approximately 0.8\% in accuracy in the dense model and 0.9\% in the sparse model, showing that the ZK framework could maintain accuracy close to the full-precision model.
Although introducing 50\% sparsity reduces the ZK accuracy by about 1.03\%, it enables significant resource savings, such as a 66.8\% reduction in memory usage as illustrated in Table~\ref{tab:framework_comparison_combined}. This balance between minor accuracy loss and substantial resource reduction highlights the effectiveness of sparsity and teleportation in optimizing models for both efficiency and verifiability.

\begin{table}[t]
\centering
\adjustbox{max width=\columnwidth}{
\begin{tabular}{@{}llccccc@{}}
\toprule
Model & Dataset & Sparsity & \begin{tabular}[c]{@{}c@{}}Full Precision \\ Model Accuracy (\%)\end{tabular} & \begin{tabular}[c]{@{}c@{}}ZK Accuracy \\ (\%)\end{tabular} \\ \midrule

\multirow{3}{*}{\centering ResNet-20} & \multirow{3}{*}{\centering CIFAR-100} 
& 0\% & 68.7 & 67.9 \\
&  & 50\% & 68.1 & 67.2 \\ \cmidrule(l){3-5}
&  & \textbf{Reduction (\%)} & 0.87\% & 1.03\% \\ \midrule

\multirow{3}{*}{\centering Tiny-ViT} & \multirow{3}{*}{\centering ImageNet} 
& 0\% & 72.2 & 71.4 \\
&  & 50\% & 71.1 & 70.7 \\ \cmidrule(l){3-5}
&  & \textbf{Reduction (\%)} & 1.1\% & 0.7\% \\

\bottomrule

\end{tabular}
}
\caption{Impact of sparsification and ZK circuit quantization on accuracy.}
\label{tab:acc}
\vspace{-12pt}
\end{table}

\subsection{Ablation Study}
\label{sec:ablation}

To mitigate the computational overhead of ZK proof generation, particularly for the prover, we incorporated sparsification and teleportation into our design. To evaluate the individual contributions of each module, we present the ablation study results in Table~\ref{tab:ablation}.

\begin{table*}[tbh]
\centering
\adjustbox{max width=\textwidth}{
\begin{tabular}{@{}llccccc@{}}
\toprule
Model & Dataset & \begin{tabular}[c]{@{}c@{}}Post-Processing \\ (Method)\end{tabular} & \begin{tabular}[c]{@{}c@{}}Memory Usage \\ (GB)\end{tabular} & \begin{tabular}[c]{@{}c@{}}Proving Time \\ (s)\end{tabular} & \begin{tabular}[c]{@{}c@{}}Verification Time \\ (ms)\end{tabular} \\ \midrule

\multirow{4}{*}{\centering ResNet-20} & \multirow{4}{*}{\centering CIFAR-100} 
& No Post-Processing & 120.2 & 564 & 3.1 \\
&  & Teleportation & 73.4 (38.9\%) & 430 (23.8\%) & 2.9 (6.5\%) \\
&  & Sparsification 50\% & 60.7 (49.5\%) & 368 (34.8\%) & 2.5 (19.4\%) \\
&  & Both Sparse and Teleported & 39.8 (66.9\%) & 307 (45.6\%) & 2.5 (19.4\%) \\ \midrule

\multirow{4}{*}{\centering MobileNetv1} & \multirow{4}{*}{\centering CIFAR-10} 
& No Post-Processing & 139.0 & 779 & 3.5 \\
&  & Teleportation & 138.3 (0.5\%) & 757 (2.8\%) & 3.4 (2.9\%) \\
&  & Sparsification 50\% & 85.3 (38.6\%) & 473 (39.3\%) & 2.8 (20.0\%) \\
&  & Both Sparse and Teleported & 56.7 (59.2\%) & 358 (54.0\%) & 2.7 (22.9\%) \\ \midrule

\multirow{4}{*}{\centering Tiny-ViT} & \multirow{4}{*}{\centering ImageNet} 
& No Post-Processing & 1002.2 & 541 & 4.2 \\
&  & Teleportation & 961.5 (4.1\%) & 421 (22.2\%) & 4.2 (0.0\%) \\
&  & Sparsification 50\% & 838.3 (16.4\%) & 310 (42.7\%) & 3.8 (9.5\%) \\
&  & Both Sparse and Teleported & 814.5 (18.7\%) & 256 (52.7\%) & 3.7 (11.9\%) \\ \bottomrule

\end{tabular}
}
\caption{Comparison between the effectiveness of each post-process on the different computation resources and both the prover and verifier. The reduction percentage is computed relative to the baseline model with no post-processing. The metrics for the Tiny-ViT model correspond only to the MLP parts, as teleportation affects only the activation functions in these parts.}
\label{tab:ablation}
\end{table*}

The initial assumption may be that the main overhead in ZKP generation stems from either the large number of constraints or the extensive range of lookup tables.
However, the results in Table~\ref{tab:ablation} show that combining teleportation and sparsification yields the most substantial improvements.
In the ResNet-20, the combination reduces memory usage by 66.9\% and proving time by 45.6\%. Each post-processing method alone has a smaller impact, highlighting the complementary benefits of combining these techniques.

Applying only sparsification reduces memory usage and proving time, particularly in ResNet-20, with reductions of 49.5\% and 34.8\%, respectively, and similar benefits for MobileNetv1. Teleportation, while providing moderate improvements, is more effective for models like ResNet-20 with more lookup arguments, reducing its memory usage by 38.9\% and proving time by 23.8\%. However, its impact on MobileNetv1 is negligible, likely because the primary bottleneck in that specific model lies in the number of constraints rather than lookup overhead.
Combining teleportation with sparsification significantly outperforms sparsification alone, further reducing memory usage by up to $34.4\%$ and proving time by $24.3\%$ compared to sparsification alone.
Verification time, being less computationally demanding sees smaller reductions across configurations. Nevertheless, The combined approach still manages modest improvements, such as 19.4\% for ResNet-20 and 22.9\% for MobileNetv1.
To demonstrate the effectiveness of the optimized teleportation, we show the distribution difference between the teleported and original model activation ranges in Appendix \ref{sec:appendix_range_distribution}.

\subsection{Granular Sparsity}
\begin{figure}[th]
    \centering
    \includegraphics[width=1\linewidth]{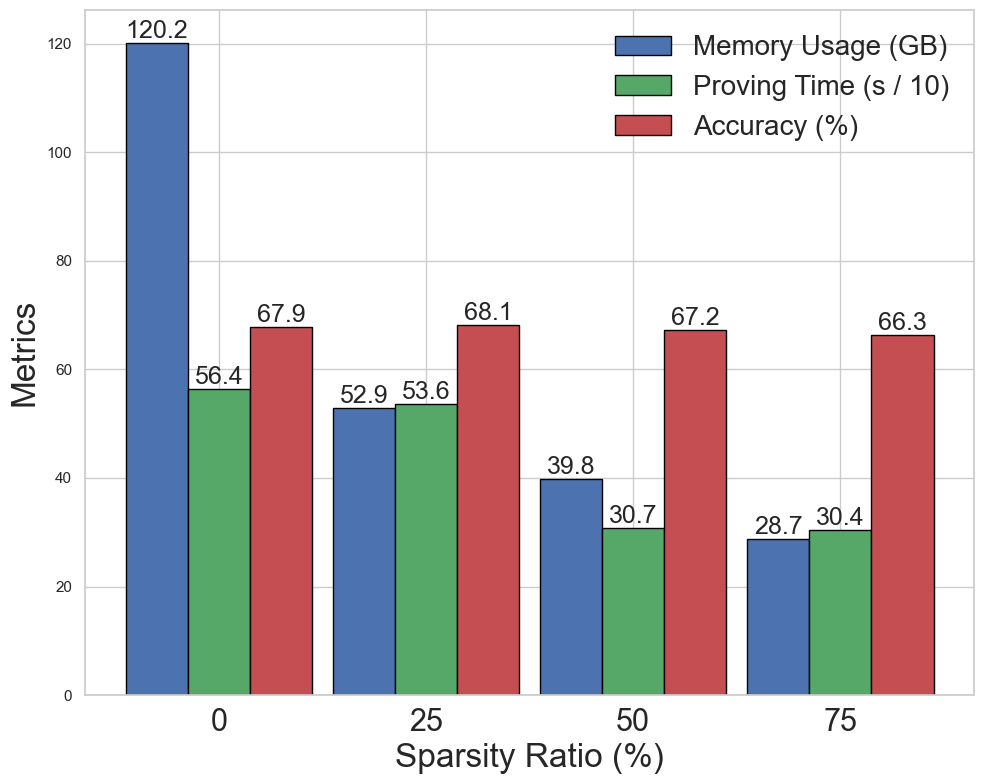}
    \caption{Effect of sparsity ratio on memory usage, proving time, and top-1 accuracy (equal to average accuracy due to balanced classes) for ResNet20 on CIFAR-100.}
    \label{fig:granular_fig}
    \vspace{-12pt}
\end{figure}

To analyze the impact of the sparsity ratio on prover computations, we conduct experiments varying the sparsity ratio from $0\%$ to $75\%$ on the ResNet model, as shown in Figure~\ref{fig:granular_fig}.

Table~\ref{tab:main_performance} demonstrates the impact of sparsity on memory usage, proving time, and accuracy for ResNet20 on CIFAR-100. Sparsification leads to a significant reduction in memory usage, with a decrease from $120.2$ GB at $0\%$ sparsity to $28.7$ GB at $75\%$. Proving time also reduces, dropping from $564$ seconds to $304$ seconds, though the reduction is less substantial compared to memory usage. Interestingly, at $25\%$ sparsity, the top-1 accuracy improves to $68.1\%$ from $67.9\%$ in the dense model, likely due to reduced quantization error as weights and activations with larger magnitudes, which are more robust to quantization, are preserved~\cite{sun2023simple,kim2023squeezellm}. However, accuracy decreases at higher sparsity levels, reaching $66.3\%$ at $75\%$. These results highlight that sparsification not only optimizes resource efficiency but can also enhance accuracy under moderate sparsity levels, though trade-offs emerge as sparsity increases.

\section{Privacy Analysis}
\label{sec:privacyAnalysis}
To conclude our analysis, we prove security of our approach against a malicious verifier that observes the model inference outputs. Our method for proving security is based on the standard secure computation framework of the real-/ideal-world paradigm. Specifically, we construct an indistinguishable (polynomial-time) simulation of the real protocol, given access only to an ideal functionality \(F\) that on model input \(X\), computes the corresponding inference output \(y\). Our proof requires that the simulator has access to the sparsity ratio of model weights, which can be observed via the sparsification process in the verification key, and is therefore a form of leakage. In Section~\ref{sec:privacy-mitigations}, we examine the possibility of removing this leakage through additional privacy measures that could form the basis of future work. 
In the following, we represent the original model weights before pruning (sparsification) as \( W^t \) and the model weights after applying the pruning algorithm as \( W_{\text{sparse}}^t \).

\subsection{Verifier View}
\label{sec:verifier-view}
The verifier has access to the neural network's input $X$, output $y$, ZK circuit embedded in the verification key (containing commitments to fixed columns $Q$ in Halo2), and the proof $\pi$ (with commitments to advice columns $A$). However, the actual weight values $W_{\text{sparse}}^t$ remain fully hidden, regardless of whether the model is sparse or dense. Moreover, each fixed column in Halo2 mixes weights, selectors, and other constants, concealing individual weight locations. Although Figure~\ref{fig:halo2_sparse_table} illustrates the weights in an ordered manner for clarity, in practice, their positions are neither predefined nor known to the verifier.

\point{Impact of sparsification on verification key} As part of the sparsification process, certain constraints are removed to reduce proving costs and improve efficiency. This affects the circuit (verification key) and thus the generated proof, as shown in Figure \ref{fig:halo2_sparse_table}, where the sparse case (B) has fewer constraints than the dense case (A). Since the verification key is visible to the verifier, they observe a commitment to  $Q_{F_0}$  in the proof and commitments to columns $A_0$, $A_1$, $A_2$  in the verification key, which shrink with sparsification. An adversarial verifier can distinguish between the verification keys of the dense and sparsified models by comparing their sizes, potentially estimating the sparsity ratio. Therefore, we must provide the sparsity ratio ($R_l$) as known leakage to the simulator algorithm when eventually constructing our proof of security. Potential measures for weakening this assumption are covered in Section~\ref{sec:privacy-mitigations}.

Note that reducing activation ranges via teleportation (Section~\ref{sec:teleportation}) does not provide additional leakage. Teleportation only restricts the input domain of neurons without altering the network function, thus not revealing additional information about the model weights.

\subsection{Proof of Privacy}
\label{sec:privacy-halo2}
We now proceed to prove the privacy of the construction against a malicious verifier, who is trying to learn the values of the model weights $W^t$. Our proof follows as a consequence of the security properties of the underlying ZK proof system.

\point{Formal privacy guarantee} The main privacy guarantee of our system is given in Theorem~\ref{theorem-simulator}; the proof follows in Appendix~\ref{sec:proof-thm-privacy}.

\begin{theorem}
\label{theorem-simulator}


For any witness \(W_{\text{sparse}}^t\) produced by the TeleSparse algorithm with sparsity ratio \(R_l\) and for any given input \(X\), there exists a simulator \(\mathcal{S}\) which, given the sparsity ratio \(R_l\), the input \(X\), and access to the ideal functionality \(F\) (without knowledge of the original witness \(W_{\text{sparse}}^t\)), constructs a simulated view \(\text{view}'\) that is indistinguishable from the real protocol view \(\text{view}\). In particular, the simulated view \(\text{view}'\) is computationally indistinguishable from the view produced during an execution of the real protocol with witness \(W_{\text{sparse}}^t\).

\end{theorem}

\subsection{Preventing Sparsification Ratio Leakage}
\label{sec:privacy-mitigations}
As previously discussed in Section \ref{sec:privacyAnalysis}, privacy loss occurs if the attacker accesses the original model's architecture and the proving system fails to protect the number of removed constraints corresponding to the pruned weights (sparsity ratio). To mitigate this privacy leakage, we propose adding dummy constraints to the circuit. The idea is to provide a noisy approximation of the number of constraints in the sparsified model. Such constraints would have no impact on the eventual computation, and thus the accuracy of the model inference procedure would go unharmed. 

On the other hand, the drawback is that the proof overhead increases proportionally with the number of added dummy constraints, and thus the proving and verification time would be computationally more expensive to execute. Furthermore, analysis of the optimal number of dummy constraints to add while still maintaining both high performance and reducing privacy leakage would require careful consideration of the noise introduced. This would likely require incorporating some notion of differential privacy into the security model, and would also require a method for allowing the PPT simulator to sample from the noisy distribution of sparsification ratios when constructing the security proof.

While the current privacy analysis provides strong justification~---~and that knowing the sparsification ratio is highly unlikely to impact privacy~---~we believe that investigating these directions further would be valuable future work. In particular, such investigation would explore the extent to which the approximation of the sparsity ratio is feasible, and to quantify how beneficial the proposed mitigation is in maintaining high performance.

\section{Conclusion and Future Work}

In this paper, we make substantial progress toward reducing the computational overhead of applying ZK-SNARKs to large-scale neural networks. By focusing on post-processing techniques, we introduce methods that improve ZK compatibility without modifying model architecture or training pipelines. Through neural network sparsification and an innovative adaptation of neural teleportation, \Name\ reduces both the number of circuit constraints and the size of lookup tables for non-linear functions. 
The method reduces prover memory usage by 67\% and proof generation time by 54\%, with minimal accuracy loss, making ZK-SNARK verification more practical for modern deep learning models.
We conducted extensive experiments demonstrating the applicability of \Name\ across various datasets and model architecture, including CNNs and transformers, to validate effectiveness in diverse settings.

Future research could extend our sparsification and activation optimization techniques for compatibility with other ZK proving systems, beyond Halo2~\cite{zcash-halo2}, thereby expanding their applicability across a broader range of ZK systems. Furthermore, although our experiments involve splitting the model and generating ZK-SNARK proofs for each part separately, we focus on selecting the split point based on the structure of transformers. However, determining the optimal split point is essential for minimizing the computational overhead of the prover. Additionally, exploring nested proof aggregation could significantly enhance scalability, especially in resource-constrained environments like edge devices.
Other paths forward include exploring methods to quantify DNN operation costs in ZK systems and better understand the interplay between sparsity, memory usage, and proof generation time for various DNN operations. This would enable more targeted sparsification, addressing the operations with the highest computational overhead in ZKP systems.
Finally, while we discuss the privacy implications of sparsification (Section~\ref{sec:privacyAnalysis}), further research into the detected leakage risk and mitigation strategies in Section~\ref{sec:privacy-mitigations} is needed.

\section{Acknowledgments}
We wish to acknowledge the thorough and useful feedback from anonymous reviewers and our shepherd. The research in this paper was supported by the UKRI  Open Plus Fellowship (EP/W005271/1 Securing the Next Billion Consumer Devices on the Edge) and EU CHIST-ERA GNNs for Network Security and Privacy (GRAPHS4SEC) projects.
Alex Davidson's work was supported by FCT through project ``ParSec: Astronomically Improving Parliamentary Cybersecurity through Collective Authorization'', ref.\ 2024.07643.IACDC, DOI \href{https://dx.doi.org/10.54499/2024.07643.IACDC}{10.54499/2024.07643.IACDC}, and the LASIGE Research Unit, ref.\ UID/00408/2025 -- LASIGE.

\bibliographystyle{ACM-Reference-Format}
\bibliography{cidarticle-example}

\appendix

\section{ZK Properties of Halo2 In the proposed System}
\label{sec:appendix_halo2_properties}
\Name, based on the Halo2 protocol, inherits its security properties including:
\begin{itemize}
    \item Succinctness: Since the generated proof is sent over the internet network, the succinctness of the generated proof is crucial to reduce the overhead of communication costs of verifiable DNN inference. The size of the Halo2 generated proof remains sub-linear relative to the complexity of $f$. Therefore, as the layers of the model grow, the size of the proof does not change drastically.
    
    \item Knowledge Soundness: A prover (computationally bounded) is unable to generate proofs for incorrect executions of $f$.
    
    \item Completeness: If the prover possesses a valid $W$ and generates the proof based on that, the proof will be successfully verified by the verifier.
    
    \item Zero-knowledge: The generated proof $\pi$ does not reveal any information about the private input ($W$) except that $y = f(X;W)$.
\end{itemize}

\section{Applications}
\label{sec:appendix_applications}
The threat model considered in this work is suitable for various real-world applications:
\begin{itemize}[leftmargin=*]
    \item \textbf{Verifiable Model Accuracy:} A model provider may want to demonstrate the accuracy of their model on a public dataset without revealing the model weights. The model provider can use ZK-SNARK to prove the accuracy of the model on the dataset, without disclosing any information about the model weights~\cite{kang2022scaling,south2024verifiable}. This ensures transparency and trustworthiness, particularly in scenarios where model reliability is critical, such as insurance. For example, insurance companies could apply ZKPs to verify the accuracy of their models in determining premiums or assessing risks~\cite{dexe2021transparency}, while keeping proprietary data (used for training the model) and model confidential.

    \item \textbf{Model Predictions as a Service:} A model provider can offer predictions as a service, where users submit input data and receive the model's output along with a ZK-SNARK proof, demonstrating that the prediction was generated correctly using the committed model weights~\cite{lu2024efficient,kang2022scaling}. This protects the provider's intellectual property while assuring users of the prediction's integrity.

    \item \textbf{Auditing ML Models:} ZK-SNARK can be employed to audit ML models for fairness and compliance with regulations~\cite{franzese2024oath}. An auditor can use ZK-SNARK to verify that a model does not discriminate against certain groups, without needing access to the model weights or the training data. This enables accountability and promotes trust in AI systems.

    \item \textbf{Proof of Ownership:} In cases of model theft or extraction, the true model owner can use ZK-SNARK to prove ownership by committing to the dataset used for training and demonstrating that their model was trained on that specific dataset. This deters theft and helps resolve ownership disputes.

    \item \textbf{Proof of Model Ownership in Generative AI:} In cases where generative AI models create content (e.g., images or text), the model provider can use ZK-SNARKs to verify that specific outputs are indeed generated by their model. By committing to the model's weights, the provider can produce a ZKP confirming that a given output, such as an image, was generated by their model. This proof of generation, without revealing model weight, reinforces ownership and helps discourage unauthorized use of generative AI models~\cite{marchal2024generative}.
\end{itemize}

\section{Notation Table}
\label{sec:appendix_notation_table}

For a comprehensive list of mathematical notations used throughout this paper, please refer to the notation table provided in Table~\ref{tab:notation_table}.

\begin{table*}[ht]
\centering

\begin{tabular}{|l|l|l|}
\hline
\textbf{Symbol} & \textbf{Description} \\
\hline
$X$ & Public input data to the DNN   \\
$y$ & Public output from the DNN computation   \\
$W$ & Weight matrix   \\
$\pi$ & Proof generated by the prover   \\
$W_{i,j}$ & Element at the $i$-th row and $j$-th column of weight matrix $W$  \\
$f(X,W)$ & The DNN computation function, taking input $X$ and weights $W$ to produce output $y$  \\
$L$ & Number of layers in the DNN   \\
$W^t$ & Original model weights before applying the sparsification algorithm     \\
$W_{sparse}$ & Sparse weight matrix after sparsification algorithm    \\
$W_{sparse}^t$ & Sparse and teleported weight matrix     \\
$L$ & Number of layers in the DNN   \\
$W^{(i)}$ & Original model weights in $i$-th layer (or block)  \\
$\widetilde{W}^{(i)}$ & Transformed (sparsified or teleported) model weights in $i$-th layer (or block)  \\

$\rho_i$ & Importance of the $i$-th weight in a block of a neural network, \\  & used in the CAP method. \\
$\delta w$ & Optimal update for the remaining weights in a block after pruning \\
$\mathbf{H}_i$ & Hessian matrix for block $i$ in a neural network, used in the CAP method. \\
$H^{-1}$ & Inverse of the Hessian matrix \\
$e_i$ & The i-th standard basis vector in $\mathbb{R}^d$, \\
 & where $d$ is the dimensionality of the vector space. \\
$R$ & Targeted sparsity ratio of the entire network in the sparsification algorithm \\

$A_{i,j}$ & Value in the $j$-th row and $i$-th advice column of the Halo2 table   \\
$F_{i,j}$ & Value in the $j$-th row and $i$-th fixed column of the Halo2 table   \\
$A$ & Set of commitments to the advice columns in Halo2 \\
$Q$ & Set of commitments to the fixed columns in Halo2 \\
$a_i(X)$ & Lagrange polynomial for the $i$-th advice column   \\
$q_i(X)$ & Lagrange polynomial for the $i$-th fixed column   \\
$Q_{F_0}(X)$ & Polynomial representing the fixed column containing model weights ($F_0$).   \\
$S_i$ & Support set for the $si$s-th row, containing indices of non-zero elements   \\
$\text{Pr}[...]$ & The probability of the event described within the brackets. \\
$\langle P^*, V \rangle(\phi)$ & The interaction between a cheating prover, $P^*$, and the verifier, $V$, on a statement $\phi$. \\ 
 & soutput is either 1 (accept) or 0 (reject). \\
$\phi$ & Statement of the ZK proof generation \\
\(\kappa\) & Security parameter \\
\(\)
$\nu(\kappa)$ & A negligible function, which approaches zero faster than any inverse polynomial. \\
$\approx$ & Computational indistinguishability \\ 
 & (two distributions are so similar that no efficient algorithm can distinguish them). \\
$S(\phi)$ & The output of a simulator $S$, an algorithm that generates a view indistinguishable \\
 & from the real view without knowing the witness \\
$ppt$ & Probabilistic Polynomial-Time   \\

$I$ & Input vector to a layer   \\
$O$ & Output vector from a layer   \\
$O_j$ & $j$-th element of the output vector $O$   \\
$\tau$ & Change of Basis (CoB) scalar   \\
$\{\tau^{(i)}\}$ & Set of CoB scalars for layer $i$   \\
$\tau^{(i)}_j$ & CoB scalar for the $j$-th neuron in the $i$-th layer   \\
$\tau^*$ & Optimal set of CoB scalars obtained through optimization.   \\
$g_j^{(i)}$ & Pre-activation input to the $j$-th neuron in layer $i$   \\
$z_j^{(i)}$ & Activation output of the $j$-th neuron in layer $i$   \\
$\phi(x)$ & Activation function in the DNN model  \\
$l(\tau)$ & Objective function for teleportation optimization   \\
$\mu$ & Small perturbation scalar used in zero-order gradient approximation methods. \\
$\lambda$ & Regularization hyperparameter balancing input range minimization and function preservation. \\

\hline
\end{tabular}
\caption{Notation Table containing symbols used in the equations}
\label{tab:notation_table}
\end{table*}

\section{Formal Halo2 and Commitment Schema Properties}
\label{sec:appendix_all_properties}
Let \(\kappa\) be a security parameter, and let \(\nu(\kappa)\) be a negligible function. The formal ZK-SNARK properties guaranteed by Halo2 are as follows:


\colonpoint{Soundness}
A proving system is \textbf{sound} if a cheating prover cannot convince the verifier of a false statement except with negligible probability.

Formally, for any efficient prover \( \mathcal{P}^* \) and any statement \( \phi \) not in the language \( L \), the probability that \( \mathcal{P}^* \) convinces the verifier \( \mathcal{V} \) to accept \( \phi \) is negligible:
\begin{equation}
\centering
\label{eq:zk_soundness}
\Pr[\langle \mathcal{P}^*, \mathcal{V} \rangle (\phi) = 1 \mid \phi \notin L] \leq \nu(\kappa)
\end{equation}

\colonpoint{Correctness}
A proving system is \textbf{correct} if an honest prover can convince the verifier of a true statement. Formally, for any statement \( \phi \in L \) and witness \( w \), the verifier accepts with overwhelming probability:

\begin{equation}
\Pr[\langle \mathcal{P}, \mathcal{V} \rangle (\phi, w) = 1] = 1 - \nu(\kappa)    
\end{equation}

\colonpoint{Zero-Knowledge}
A proving system is \textbf{zero-knowledge} if the verifier learns nothing beyond the validity of the statement. Formally, for every probabilistic polynomial-time verifier \( \mathcal{V}^* \), there exists a simulator \( \mathcal{S} \) such that:

\begin{equation}
\text{View}(\mathcal{P}(w),\mathcal{V}^{*}(\phi)) \approx \mathcal{S}(\phi)
\end{equation}

Polynomial commitment schemes, such as KZG commitments, are cryptographic primitives used to ensure the integrity and privacy of polynomial evaluations in ZKP systems. These schemes satisfy two crucial security properties: \emph{binding} and \emph{hiding}.

\colonpoint{Binding Property}
A commitment scheme is \textbf{binding} if it is computationally infeasible for the prover to open a commitment to two different values.
Formally, let \( \text{Comm}: \mathbb{F}^n \rightarrow \mathcal{C} \) be a commitment function. The binding property ensures that for any \( \mathbf{v}, \mathbf{v'} \in \mathbb{F}^n \) with \( \mathbf{v} \neq \mathbf{v'} \), it is computationally infeasible to find \( r, r' \) such that \( \text{Comm}(\mathbf{v}; r) = \text{Comm}(\mathbf{v'}; r') \).

\begin{multline}
\label{eq:binding_property}
\forall \text{ PPT algorithms } \mathcal{A}, \quad 
\Pr\bigg[ (c, \mathbf{v}, \mathbf{v'}) \leftarrow \mathcal{A}() \, \big| \\
\operatorname{Comm}(\mathbf{v}; r) = c = \operatorname{Comm}(\mathbf{v'}; r') 
\wedge \mathbf{v} \neq \mathbf{v'} \bigg] \leq \nu(\kappa),
\end{multline}

\colonpoint{Hiding Property}
A commitment scheme is \emph{hiding} if the committed value remains indistinguishable to any adversary without knowledge of the opening randomness. Formally, for any \( \mathbf{v}, \mathbf{v'} \in \mathbb{F}^n \) and independent randomness \( r, r' \in \mathbb{R} \), the distributions of \( \operatorname{Comm}(\mathbf{v}; r) \) and \( \operatorname{Comm}(\mathbf{v'}; r') \) are computationally indistinguishable. Specifically, for any PPT adversary \( \mathcal{A} \), it holds that:

\begin{equation}
\label{eq:commitment-hiding}
\left| 
\Pr\big[\mathcal{A}(\operatorname{Comm}(\mathbf{v}; r)) = 1\big] - 
\Pr\big[\mathcal{A}(\operatorname{Comm}(\mathbf{v'}; r')) = 1\big]
\right| \leq \nu(\kappa)
\end{equation}

\section{Formal Proof of Sparsification Soundness}
\label{sec:appendix_formal_proof}
We aim to prove that excluding constraints for zero-valued entries does not allow a cheating prover to convince the verifier of a false statement, i.e., that \( \mathbf{O'} \neq W I \). This proof relies on the formal properties of Halo2 and commitments, as detailed in Appendix~\ref{sec:appendix_all_properties}.

Define support for each \( i \):
\[
S_i = \{ j \in \{1, \dots, d_2\} \mid W_{i,j} \neq 0 \}
\]
The constraint for each \( i \) should represent the following equation:
\[
O_i = \sum_{j \in S_i} W_{i,j} I_j
\]

Based on the circuit representation in equation~\ref{eq:table_constraints_linearlayer}, the rows which $F_{0i}$ is zero will be discarded as shown in Figure~\ref{fig:halo2_sparse_table} so the polynomials $A_0 , Q_{F_0} , A_1 , A_2$ in equation~\ref{eq:polynomial_constraints_linearlayer} only interpolate remained cells. Consequently the commitments is computed based on the new polynomial functions. In other words, in the verification key only the commitments to the non-zero weights exist. 

To prove the soundness, we consider all possible cheating strategies by the prover to somehow change the output of the linear layer ($O_i$).

\colonpoint{Step 1, Altering the Matrix \( W \)}  
Assume we construct an adversary \( \mathcal{A} \) against the binding property of the commitment scheme, utilizing an adversary \( \mathcal{B} \) that attempts to alter \( W \) while generating a valid proof (accepted by the verifier).  
Since \( W \) is stored in fixed column \( F_0 \), and commitments \( \text{Commit}(Q_{F_0}(X)) \) are included in the verification key, any attempt to alter \( W \) by adversary \( \mathcal{B} \) would require finding a new matrix \( W' \) such that  
\[
\text{Commit}(W, r) = \text{Commit}(W', r').
\]  
If adversary \( \mathcal{B} \) could successfully alter \( W \) by finding \( W' \) and \( r' \), adversary \( \mathcal{A} \) could utilize \( \mathcal{B} \) to break the binding property of the commitment scheme. This contradicts the assumption that the commitment scheme is secure under standard cryptographic assumptions.  
Therefore, \( \mathcal{B} \) cannot alter \( W \).

\colonpoint{Step 2, Altering the Witness Vector \( \mathbf{I} \)}  
Given that \( W \) cannot be altered, the only remaining strategy for adversary \( \mathcal{B} \) is to modify the witness vector \( \mathbf{I} \) to \( \mathbf{I}' \) in a way that affects the output \( \mathbf{O} \).  
Again, we construct an adversary \( \mathcal{A} \) against the soundness of the copy constraints in Halo2. Adversary \( \mathcal{A} \) utilizes \( \mathcal{B} \), which attempts to modify the witness vector \( \mathbf{I} \) to \( \mathbf{I}' \).  

However, the circuit copy constraints enforce consistency between layer inputs and outputs. These constraints ensure that the input \( I_j \) of the current layer \( l \) must exactly match the output \( O_j \) from the previous layer \( l-1 \). 
If adversary \( \mathcal{B} \) attempts to modify the values of \( \mathbf{I} \) corresponding to zero entries in the matrix \( W \), this will not affect the output \( \mathbf{O} \), as those terms contribute nothing to the summation in Equation~\eqref{eq:matrix_vector_mul}. Thus, the adversary cannot achieve an invalid output \( O_{i}' \) through these modifications. To have any effect, adversary \( \mathcal{B} \) attempt must involve changing values in \( \mathbf{I} \) that correspond to the non-zero entries in \( W \).
Assume that adversary \( \mathcal{B} \) successfully modifies \( I_j \) to \( I_j' \) corresponding to non-zero entries (modifies a value \( I_j \) where \( j \in S_i \)) and then generates a valid proof based on this. Therefore, adversary \( \mathcal{A} \) could exploit adversary \( \mathcal{B} \) to construct a strategy to break the copy constraints of Halo2. This would lead to a contradiction, as the copy constraints of Halo2 are enforced by the soundness property of the ZK-SNARK system.  
Consequently, \( \mathcal{B} \) cannot successfully alter \( \mathbf{I} \) without being detected by the verifier.

\colonpoint{Conclusion}  
If adversary \( \mathcal{B} \) cannot alter \( W \) due to the binding property of the commitment scheme, and cannot modify \( \mathbf{I} \) without violating the copy constraints, then no cheating strategy remains.  
As a result, eliminating constraints for zero entries in fixed columns does not compromise the soundness of the proving system.

\section{Proof Splitting}
\label{sec:appendix_splitting}

To ensures that the ZK-SNARK system remains scalable for use on devices with limited resources, we could leverage Halo2's recursive proof composition capabilities, allowing complex proofs to be split and incrementally verified. The splitting not only reduces the load on constrained devices but also enhances the overall scalability of the ZK-SNARK verification process.

\begin{figure}
    \centering
    \includegraphics[width=1\linewidth]{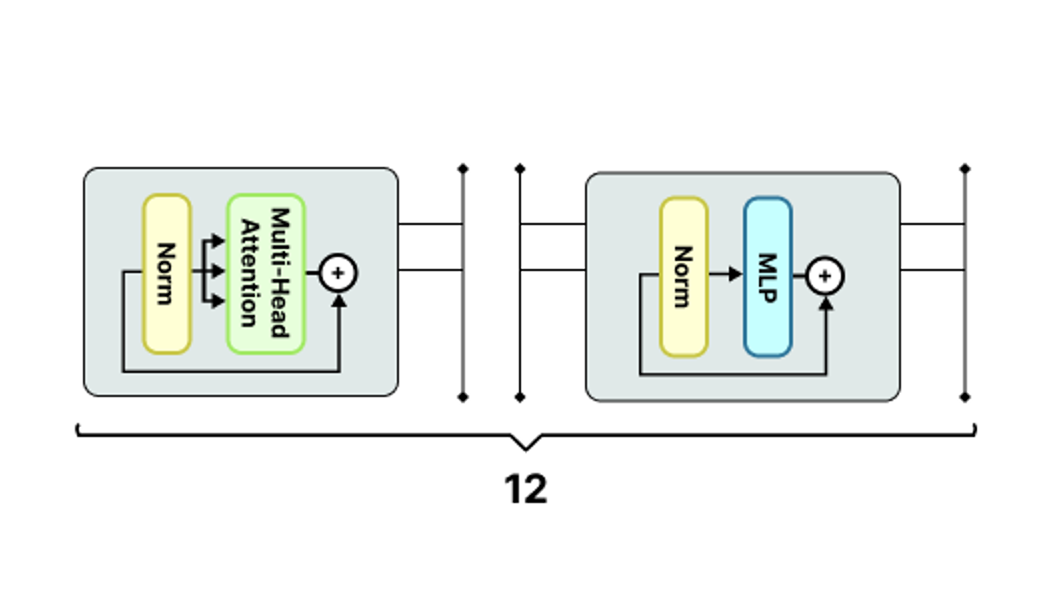}
    \caption{Transformer architecture contains two parts of attention and MLP. We split each part to generate proof in a separate ZK circuits.}
    \label{fig:vit_splitting}
\end{figure}

The total number of constraints regarding a model like a vision transformer is significantly huge. 
However, by splitting the layers of a DNN (e.g., ViT shown in Figure~\ref{fig:vit_splitting}), each part could be proved separately. 
In particular, we could split a ViT into our desired $M$ parts such that the output of the $(i-1)$-th part is the input of the $i$-th part. 
It is easy to do that in most DNN models including ViT, which consists of 12 layers, and each layer has two main components: self-attention and MLP. 
For instance, given a sparse ViT, we split the model into $M = 24$ parts, which are non-overlapping. 
For each part, the corresponding proving key $pk_i$ and the corresponding verification key $vk_i$ are generated by halo2. 
Also, we should contain the commitment to the output of the $(i-1)$-th part as the public input to the circuit corresponding to the $i$-th part. 
Proof generation for each partition can proceed either in parallel or sequentially, with specific advantages in each approach that impact the scalability of ZK-SNARKs in deep learning contexts. In a parallel setup, proof generation time remains fixed, regardless of the number of partitions  M , as each segment can be processed concurrently without increasing overall latency. This parallel configuration is especially beneficial in distributed systems, where multiple nodes can simultaneously handle each partition's proof, thus minimizing total proving time for large models.
Conversely, in a sequential setup, memory requirements remain constant and independent of  $M$ , as each partition can be processed one at a time with minimal memory overhead. This sequential approach is particularly advantageous for memory-constrained environments, where a single proof can be generated step-by-step through the partitions, thus avoiding the exponential memory growth that would typically accompany large models in ZK-SNARK-based proofs for deep networks. This flexibility in proof generation pathways ensures scalability and practicality of ZK-SNARKs for complex architectures like ViTs, where constraint management is critical. Detailed performance measurements of this partitioning approach, specifically on the Vision Transformer (ViT), are provided in the Table~\ref{tab:main_performance}.

In the verification step, the verification of the $i$-th part, the verifier should check that the committed output of the $(i-1)$-th part is equal to the commitment of the input of the $i$-th part.

Building on this partitioned proof generation, Halo2 enables proof composition via Nested Amortization~\cite{bowe2019recursive} and an Accumulation Scheme~\cite{zcash-halo2}, allowing individual proofs from each partition to be combined into a single recursive proof with constant verification cost. This makes the proof size independent of the partition count. This means that only a single verification check is required, leveraging Halo2's recursive ZK-SNARKs to verify all parts efficiently. In verification, a single recursive check confirms that the final proof upholds commitments across all partitions $1$ to $M$, implicitly verifying alignment of inputs and outputs between partitions. Proof Composition in Halo2 is completely beneficial for DNNs by keeping the verification cost and proof size constant with respect to $M$. Proof aggregation is particularly advantageous for applications like on-chain verification, where constant proof size and verification cost significantly reduce computational overhead. However, our experiments focused on partitioned proof generation and generating proof for each part separately, excluding recursive composition, as it lies beyond the scope of this work.

\section{Effect of Teleportation on Activation Range Distribution}
\label{sec:appendix_range_distribution}

\begin{figure}
    \centering
    \includegraphics[width=1\linewidth]{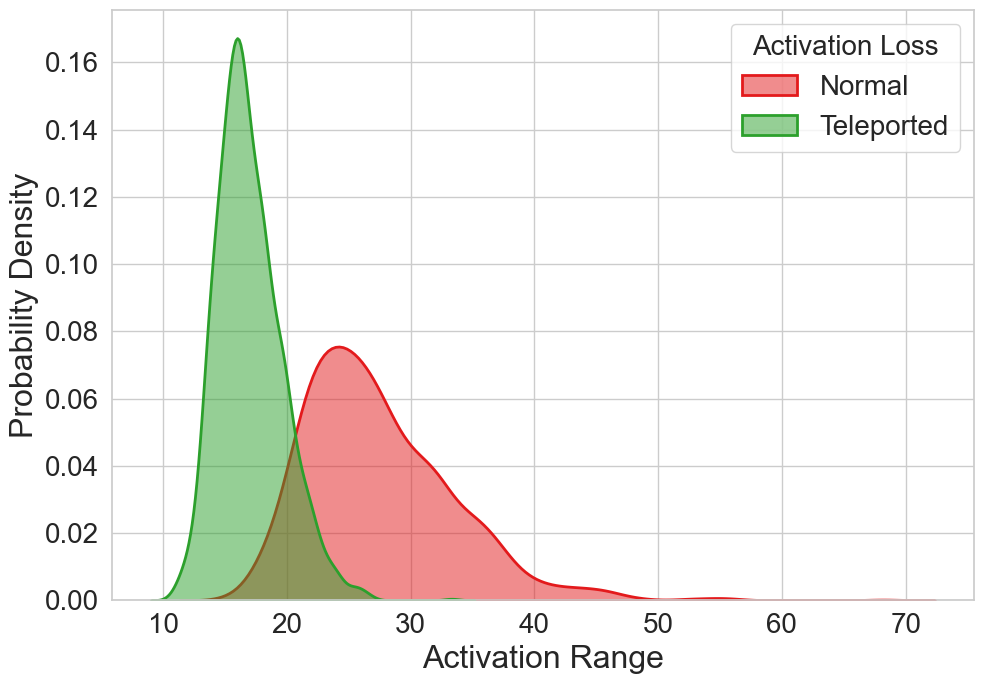}
    \caption{Activation range distribution of 300 CIFAR100 samples on ResNet-20: Red represents the original model's activation range with outliers, and green shows the distribution after optimized teleportation}
    \label{fig:range_distribution_resnet20}
\end{figure}

To demonstrate the effectiveness of the proposed optimized teleportation in reducing activation function range values, as discussed in Section \ref{sec:teleportation} and the experiments in Table \ref{tab:ablation}, we present the distribution of activation ranges for $300$ CIFAR-100 samples on ResNet-20 in Figure \ref{fig:range_distribution_resnet20}. The original network's activation range is long-tailed, with some outliers exceeding a range of $70$. After applying teleportation, the distribution narrows considerably, with reduced variance without extreme values, indicating the success of the teleportation in minimizing the activation range. 
Specifically, the Mean Activation Loss is reduced from 27.39 (Original) to 16.98 (Teleported), showing a reduction of approximately 38.01\%. Similarly, the Standard Deviation of Activation Loss decreases from 5.99 (Original) to 2.64 (Teleported), a reduction of 55.9\%, indicating a more compact and consistent distribution after teleportation.
As mentioned in Table \ref{tab:ablation}, eliminating outlier values is crucial for reducing the prover's resource overhead. Outliers can significantly increase the complexity of the lookup constraints in ZK-SNARK proof generation, leading to higher CPU and RAM costs. By minimizing these outliers, the activation range distribution becomes more compact, thus optimizing the prover's resource utilization and improving the efficiency of the verification process.

\section{Proof of Theorem~\ref{theorem-simulator}}
\label{sec:proof-thm-privacy}
To prove Theorem~\ref{theorem-simulator}, we must show that the simulator \( \mathcal{S} \) can reconstruct all elements of \( \text{view}' \) such that the verifier cannot distinguish \( \text{view}' \) from \( \text{view} \). Let \(\kappa\) be the security parameter, and let \(\poly(\kappa)\) and \(\negl(\kappa)\) denote polynomial and negligible functions, respectively.
As defined in Section \ref{sec:verifier-view}, the simulated verifier view is: \(\text{view}' = (\pi', X, y, vk', R_l)\), where the verification key \(vk'\) contains commitments to the fixed columns represented by \(Q\) (as shown in Equation~\eqref{eq:verificationkey}).
Recall that the simulator has access to an ideal functionality \( F \), which, given the input data \( X \) sent by the verifier, outputs the corresponding label \( y \).
The simulator does not have access to the true model weights \( W_\text{sparse}^t \) but can generate model weights, denoted as $V^t_\text{sparse}$, that produce \( y \) given \( X \), with the weights randomly pruned according to the given layer-wise sparsity ratio $R_l$.

The simulator \( \mathcal{S} \) constructs \( \text{view}' \) as following steps:
\begin{enumerate}
    \item \textit{Verification Key Simulation:}  
    The simulator first generates a verification key \( vk' \) by committing to model weights \( V_{\text{sparse}}^{t} \) following the defined sparsity ratio, without any knowledge of \( W_i^t \).
    The simulator then sends \( vk' \) to the verifier.
    For simplicity, assume that the commitment of weights is performed layer-wise. Specifically, for layer \( l \), the commitment to the polynomial \( f_l(X) \), which interpolates the remaining weights \( V_{\text{sparse}}^{t,(l)} \), is included in the verification key.
    The position of each weight is not included in the commitment as only the commitment is made to the polynomial \( f_l(X) \) of degree \( \|V_{\text{sparse}}^{t,(l)}\|_0 - 1 \), derived via Lagrange interpolation of the remaining weights.
    Since the number of values in \( V_{\text{sparse}}^{t,(l)} \) matches those in \( W_{\text{sparse}}^{t,(l)} \), resulting in equal polynomial degrees, and given the hiding property of the commitment scheme (Equation~\eqref{eq:commitment-hiding}), the probability that the verifier distinguishes between the simulator and TeleSparse weight commitments are less than \(\nu(\kappa)\), demonstrating the computational indistinguishability of the verification keys corresponding to the two weights.

    \item \textit{Output Simulation:}  
    Given the input data \( X \) provided by the adversarial verifier, the simulator uses the ideal functionality \( F \) to compute the corresponding output label \( y \). Therefore, $X$ and $y$ are identical in $view$ and $view'$.

    \item \textit{Proof Simulation:}
    Using the simulated proof generation procedure (afforded by the zero-knowledge property of the NIZK proof system):
    \[
    \text{SimProve}(\text{trap}, X, y, vk') \to \pi'
    \]
    the simulator generates the proof \( \pi' \) and provides is to the verifier. Since \( \pi' \) is a simulated proof constructed using the trapdoor \( \text{trap} \), it is computationally indistinguishable from a real proof \( \pi \). This follows from the fact that Halo2 (the underlying ZK proof system) is a valid NIZK proof system~\cite{zcash-halo2} (via the zero-knowledge property, Appendix~\ref{sec:appendix_all_properties}). Notably, as the sparsity ratio of $V^t_\text{sparse}$ is the same as $W_\text{sparse}^t$ in each layer, the number of constraints in both cases is exactly equal, resulting in the same proof size.
    \qed
\end{enumerate}

\end{document}